\documentclass[twocolumn,amsthm]{autart}

\counterwithin*{section}{part}

\usepackage{xcolor}
\usepackage{textcomp}
\usepackage[authoryear,round]{natbib}
\usepackage{amsmath,amssymb,amsfonts,amsthm}
\usepackage{graphicx}
\usepackage{wrapfig}
\usepackage{array,booktabs,tabularx,multirow}
\usepackage{bbm}
\usepackage{algorithm,algorithmic}
\usepackage{tikz}
\usepackage{nicematrix}
\usetikzlibrary{automata,positioning,shapes,backgrounds,shapes.geometric,decorations.pathreplacing}
\usepackage{adjustbox}

\usepackage[colorlinks=true,citecolor=blue,linkcolor=blue,urlcolor=blue]{hyperref}
\usepackage[capitalize]{cleveref}
\crefname{assumption}{Assumption}{Assumptions}
\crefname{theorem}{Theorem}{Theorems}

\newtheorem{theorem}{Theorem}

\newtheorem{lemma}{Lemma}
\newtheorem{corollary}{Corollary}

\newtheorem{assumption}{Assumption}
\newtheorem{remark}{Remark}

\begin{document}
\begin{frontmatter}
\title{Analysis of On-Policy Policy Gradient Methods Under Distribution Mismatch}
\author[sjtu]{Weizhen Wang}\ead{wangwz.im@gmail.com},
\author[sjtu]{Xinzhe Zhou}\ead{zhou\_xinzhe@sjtu.edu.cn},
\author[sjtu]{Jianping He}\ead{jphe@sjtu.edu.cn},
\author[sjtu]{Xiaoming Duan\thanksref{corresponding}}\ead{xduan@sjtu.edu.cn}
\thanks[corresponding]{Corresponding author: Xiaoming Duan.}
\address[sjtu]{School of Automation and Intelligent Sensing, Shanghai Jiao Tong University, and Key Laboratory of System Control and Information Processing, Ministry of Education of China, Shanghai, China}
\begin{abstract}

		Policy gradient methods are among the most successful approaches for  solving challenging reinforcement learning problems. Despite their empirical success, many state-of-the-art policy gradient algorithms for discounted problems estimate update directions using a sampling distribution that differs from the discounted state distribution required by the policy gradient theorem. In this work, we analyze the impact of this distribution mismatch on  policy gradient methods. Specifically, {for tabular direct and softmax parameterizations, we establish global convergence of objective values under biased updates}.  {For tabular softmax policies, we further derive the exact leading-order coefficients of the asymptotic objective errors under biased and normalized exact updates. Their ratio provides an explicit criterion for determining when distribution mismatch accelerates or slows asymptotic convergence.} We then extend the  analysis to more general policy parameterizations by deriving explicit bounds on both the state distribution mismatch and the induced normalized policy-gradient mismatch in episodic and continuing MDPs. These mismatches vanish at least linearly as the discount factor approaches one. 
		Under an additional smoothness assumption, we further show that the limit inferior of the squared norm of the normalized exact policy-gradient direction along the biased iterates is bounded by $O((1-\gamma)^2)$ in both settings. Our results provide theoretical  insights into the robustness of policy gradient methods to distribution mismatch and shed light on the gap between their theoretical formulations and practical implementations.
	
\end{abstract}
\end{frontmatter}
\section{Introduction}
	\label{sec:introduction}
	\subsection{Problem Description and Motivation}
	{Policy} gradient  algorithms aim to identify a  parameterized policy that  optimizes a performance metric by iteratively updating the policy parameters along a  gradient direction. In reinforcement learning (RL), popular policy gradient  methods, such as  Trust Region Policy Optimization (TRPO)~\citep{JS-SL-PM:2015}, Asynchronous Advantage Actor-Critic (A3C)~\citep{VM-AB-KK:2016}, and Proximal Policy Optimization (PPO)~\citep{JS-FW-PD:2017} are widely used to maximize the \textcolor{black}{expected} discounted cumulative reward and  have demonstrated strong empirical performance.  Since the policy gradient can be expressed as an expectation with respect to a state distribution, it is commonly
    estimated from trajectory samples in model-free implementations. However, as noted in~\citep{PT:2014}, there exists a discrepancy between the practical sampling distribution and the \textcolor{black}{discounted state distribution required by the policy gradient theorem}. Specifically,\textcolor{black}{many on-policy implementations} \textcolor{black}{collect state-action pairs by executing} the current policy, \textcolor{black}{store them in a rollout buffer, and estimate the update direction using minibatches sampled uniformly from the buffer}. 
	Since the theoretically required state distribution \textcolor{black}{weights state visits according to their temporal discount factors}, uniform sampling from the rollout buffer \textcolor{black}{without corresponding weighting or correction} leads to a distribution mismatch.

	This distribution mismatch introduces bias into gradient estimates, resulting in biased update directions in many widely used policy gradient algorithms. 
	\textcolor{black}{Consequently, standard guarantees for exact gradient ascent do not directly apply, and such biased updates can lead to suboptimal performance in certain settings}~\citep{CN-PT:2020}.
	Nevertheless, due to their ease of implementation and strong empirical performance, biased policy gradient  algorithms remain widely adopted, \textcolor{black}{often without explicitly correcting the distribution mismatch}.
	\textcolor{black}{Understanding the effect of this mismatch and why such algorithms can remain effective therefore calls for further theoretical analysis.}.

	\subsection{Related work}
	The state distribution mismatch in policy gradient methods has been highlighted in numerous studies \citep{PT:2014,CN-PT:2020,SZ-RL-VH:2022,SW-LS-GT:2022,ST-AP-RM:2022,FC-MG-AM:2023,HP-DY-MS:2023}.
	\textcolor{black}{Existing studies have investigated both its effects on policy optimization and methods for correcting the resulting bias.}
	Zhang et al.~\citeyearpar{SZ-RL-VH:2022} consider this mismatch in the \textit{episodic} setting.
	\textcolor{black}{Using the temporal policy-gradient formulation in~\citep{CN-PT:2020}, they implement DisPPO by restoring the factor $\gamma^t$ in the actor update for each transition observed at time $t$ within an episode.}
	This approach requires \textcolor{black}{tracking the within-episode time index and assigns exponentially smaller weights to transitions occurring later in an episode}.
	\textcolor{black}{Their experiments show that the uncorrected PPO can outperform DisPPO in some tasks, and they interpret this advantage through the role of later transitions in representation learning using an auxiliary-task formulation.}
	Since the bias  originates from  a distribution shift,  another common approach is to correct the mismatch through  importance sampling.   Che~et~al.~\citeyearpar{FC-MG-AM:2023} focus on the \textit{continuing} setting and introduce an averaging correction factor that theoretically corresponds to the importance ratio.
	To approximate this ratio in practice, they collect \textcolor{black}{timestamped transitions in a rollout buffer and train an auxiliary neural network by regression}. 
	\textcolor{black}{This state-dependent correction provides a lower-variance alternative to directly weighting samples by $\gamma^t$, but requires learning an additional correction model.}
	
	\begin{sloppypar}
Instead of correcting the distribution, Tosatto~et~al.~\citeyearpar{ST-AP-RM:2022} derive an alternative formulation of policy gradient that avoids the discounted distribution by connecting it with the gradient of the action value function $\nabla_\theta Q^\pi$, which they  refer to as the gradient critic.  If both the action value function $Q^\pi$ and the gradient critic $\nabla_\theta Q^\pi$ are known, then an exact policy gradient can be computed  by averaging over the initial state distribution,  thereby avoiding reliance on the discounted state distribution.  They further show that the gradient critic satisfies a Bellman equation, allowing it to be estimated  through temporal-difference methods. \textcolor{black}{This approach avoids explicit distribution correction but requires estimating the gradient critic in addition to the action-value function.} Pan~et~al.~\citeyearpar{HP-DY-MS:2023} investigate several  heuristic bias rectification approaches, \textcolor{black}{including learning rates adjustment, adaptive optimizers, and KL regularization},  through extensive experiments, but their study
    is primarily empirical.
\end{sloppypar}

	\textcolor{black}{ These studies provide methods for correcting the mismatch and insights into the empirical effectiveness of biased updates.
		Complementing these efforts, we focus on the optimization guarantees that remain available without distribution correction.
		Specifically, we study whether biased updates preserve global convergence under tabular parameterizations, {compare the exact asymptotic convergence rates of biased and  exact updates under tabular softmax parameterization}, and how the mismatch affects stationarity guarantees under more general parameterizations.
	}
	
	\subsection{Contributions}
	Our main contributions are summarized as follows:
	\begin{itemize}
		\item \textbf{Convergence and exact asymptotic rates under tabular parameterizations.} We establish global objective-value convergence under biased updates for both direct and tabular softmax parameterizations. {For tabular softmax policies under a unique optimal action assumption, we further derive the exact leading-order coefficients of the asymptotic objective errors for the biased and      exact updates. Their ratio provides an explicit comparison of their asymptotic convergence speeds.}
		\item  \textbf{Distribution and gradient mismatch bounds  for general parameterizations.} 
		\textcolor{black}{For episodic and continuing MDPs, we derive explicit bounds on the mismatch between the discounted and undiscounted state
        distributions. Under general policy parameterizations with bounded policy derivatives, we further bound the induced normalized policy-gradient mismatch and show that it vanishes at least linearly as $\gamma\to1$.}
		\item \textbf{Optimization guarantees under objective smoothness.} {Under an additional smoothness assumption on the objective, we bound the limit inferior of the squared norm of the normalized exact policy-gradient direction along the biased iterates by $O((1-\gamma)^2)$ in both episodic and continuing MDPs.}
\end{itemize}

A preliminary  conference version of this work has been accepted for publication in~\citep{WW-JH-XD:2026CDC}. This paper extends that version with exact asymptotic objective-error coefficients, stronger tabular convergence results allowing arbitrary positive constant stepsizes, and sharper mismatch and approximate-stationarity bounds, improving the continuing-MDP guarantee from $O(1-\gamma)$ to $O((1-\gamma)^2)$. It also provides complete proofs and expanded experiments validating the asymptotic coefficients and comparing neural-network-based actor--critic methods.

\subsection{Organization} The rest of the paper is organized as follows. \cref{sec:background} introduces policy gradient methods and the state distribution mismatch considered in this work.
\cref{sec:analysis} establishes global convergence under
biased tabular updates, derives exact asymptotic rates
for tabular softmax policies, and presents mismatch bounds
and optimization guarantees for general parameterizations.
\cref{sec:simulation} provides numerical experiments
illustrating the theoretical results in episodic and
continuing tasks.
\cref{sec:conclusion} concludes the paper.

\section{Background} \label{sec:background}
\subsection{Policy gradient methods}
\textbf{Markov decision processes:} We consider a finite  Markov decision process (MDP) $M=\langle\mathcal{S}, \mathcal{A}, R, \mathbb{P}, \gamma,  d_0\rangle$, where $\mathcal{S}$ and $\mathcal{A}$ are the \textcolor{black}{finite} state  and action spaces, respectively, $R:\mathcal S\times\mathcal A\to\mathbb R$ is the reward function, $\mathbb{P}\left(s^{\prime}  \, | \,  s, a\right)$ is the of transitioning to state $s'$  from state  $s$ after taking action $a$,  $\gamma \in[0,1)$ is the discount factor, and $d_0$ is the initial state distribution. We assume that rewards are uniformly bounded:  $|R(s,a)| \leq R_{\max}$ for all $s\in\mathcal{S}$ and $a\in\mathcal{A}$.

\textbf{Policy and performance metric:} A \textcolor{black}{stationary}  policy $\pi: \mathcal{S} \times \mathcal{A} \to[0,1]$ satisfies $\sum_{a\in\mathcal{A}}\pi(a\,|\,s)=1$ for all $s\in\mathcal{S}$,  where $\pi(a  \, | \,  s)$  denotes the probability of selecting action $a$ at state $s$.
\textcolor{black}{We seek to maximize the expected discounted return}
$  J(\pi) \triangleq \mathbb{E} _\pi \left[ \sum_{t=0}^\infty \gamma^t R_{t} \right]$, where $R_t=R(s_t,a_t)$ and the expectation is taken over trajectories starting from $s_0\sim d_0$ and following policy $\pi$.
The state value function and the state-action value function are defined,
respectively, as $V^\pi(s)\triangleq \mathbb{E} _\pi \left[ \sum_{t=0}^\infty \gamma^t R_{t} \,|\, s_0=s \right]$ and $Q^\pi(s,a) \triangleq \mathbb{E} _\pi \left[ \sum_{t=0}^\infty \gamma^t R_{t} \,|\,  s_0=s,a_0=a \right]$. 
The objective can be written as $J(\pi)=\sum_{s\in\mathcal S}d_0(s)V^\pi(s)$~\citep{SZ:2025}. For a fixed stationary policy $\pi$,  the induced Markov chain has transition matrix $P^\pi$ given by  $P^\pi(s,s')= \sum_{a\in\mathcal{A}} \pi(a\, | \, s)\mathbb{P}(s' \, | \, s,a)$.

\textbf{Continuing and episodic MDPs:} We consider both continuing tasks, in which the process evolves indefinitely, and episodic tasks, in which each episode terminates upon reaching a terminal state.
{Without loss of generality, we  represent an episodic MDP using a single absorbing state $z$, since multiple terminal states can be merged without changing the returns. At $z$,  $\mathbb{P}(z \, | \, z,\cdot)=1$ and  $R(z,\cdot)=0$, and hence $V^\pi(z)=0$.}
Since actions at the absorbing state do not affect the return, we define the set of states relevant to policy
optimization as $\tilde{\mathcal S} $, which satisfies $\tilde{\mathcal S} =\mathcal S$
in continuing MDPs and
$\tilde{\mathcal S} =\mathcal S\setminus\{z\}$
in episodic MDPs.
We assume that $d_0(s)>0$ for every
$s\in\tilde{\mathcal S} $, following similar assumptions in \citep{JM-CX-DS:2020,AA-SK-GM:2021}.
For episodic MDPs, we additionally assume $d_0(z)=0$. This entails no loss of generality provided $d_0(z)<1$: removing the initial probability mass at $z$ and renormalizing the remaining distribution only rescales
$J(\pi)$ by a positive constant  since $V^\pi(z)=0$, and therefore leaves the optimal policies  unchanged.

\textbf{Policy gradient:} We represent the policy by a parameterized function $\pi_\theta$ and write $J(\theta)$ as shorthand for $J(\pi_{\theta})$. Gradient-based methods iteratively update the parameters according to $\theta_{t+1}=\theta_t+\eta_t \nabla_\theta J\left(\theta_t\right)$, where $\eta_t$ is the step size. \textcolor{black}{For brevity, we write $\pi$ for $\pi_\theta$. The policy gradient theorem}~\citep{RS-DM-SS:1999} gives
\begin{equation} \label{eq:gradientJ}
	\nabla_\theta J (\theta) =   \sum_{s \in \mathcal{S}} \mu_{\pi,\gamma}(s)  \sum_{a\in \mathcal{A}} \nabla_\theta \pi(a \, | \,  s) Q^\pi (s,a),
\end{equation}
where $\mu_{\pi,\gamma}$ is the discounted state occupancy measure  of $\pi$:
\begin{equation} \label{eq:mu}
	\mu_{\pi,\gamma}(s) = \sum_{t=0}^\infty \gamma^t \mathbb{E} _{s_0 \sim d_0}  \left[\mathbb{P}(s_t = s  \, | \,  s_0, \pi) \right].
\end{equation}

\subsection{Distribution mismatch}
Although \eqref{eq:gradientJ} provides an exact characterization of the policy gradient, it is generally not directly computable when the MDP model is unknown. In particular, the discounted occupancy measure \(\mu_{\pi,\gamma}\) is typically unavailable, and explicit summation  over all states  becomes impractical in large state spaces. This motivates an expectation-based representation. Specifically, let
\begin{equation}
\label{eq:kappa}
\kappa_\pi
\triangleq
\sum_{x\in\tilde{\mathcal S} }
\mu_{\pi,\gamma}(x),
\qquad
d_{\pi,\gamma}(s)
=
\frac{\mu_{\pi,\gamma}(s)}{\kappa_\pi},
\quad
s\in\tilde{\mathcal S} ,
\end{equation}
with $d_{\pi,\gamma}(z)=0$ in episodic MDPs.
Thus, $d_{\pi,\gamma}$ is a probability distribution on $\tilde{\mathcal S} $ obtained
by normalizing  the discounted occupancy measure $\mu_{\pi,\gamma}$.
In continuing MDPs, normalization is performed over the entire state
space and $\kappa_\pi=1/(1-\gamma)$. In episodic MDPs, we instead normalize the discounted occupancy over
the nonterminal states. This is consistent with episodic
data collection, in which the trajectory terminates upon reaching
$z$, and is also theoretically valid because
the absorbing state \(z\) with \(Q^\pi(z,\cdot)=0\) does not contribute to the policy gradient direction. We define the \emph{normalized exact policy-gradient direction} by
\begin{equation}
\label{eq:epJ}
\nabla_\theta^{\textup{e}} J \triangleq
\mathbb E_{s\sim d_{\pi,\gamma}}
\left[
\sum_{a\in\mathcal A}
\nabla_\theta\pi(a\mid s)Q^\pi(s,a)
\right].
\end{equation}
By \eqref{eq:gradientJ}--\eqref{eq:kappa}, $\nabla_\theta J =
\kappa_\pi\nabla_\theta^{\textup{e}} J $. Since $\kappa_\pi>0$, the normalization preserves the ascent
direction. In continuing MDPs, $\kappa_\pi=1/(1-\gamma)$, whereas
in episodic MDPs it may depend on the policy. Although this positive
factor can in principle be absorbed into the stepsize
\citep{RS-AB:2018}, we explicitly account for it in the subsequent
analysis.

Practical implementations, however, do not generally sample states
according to $d_{\pi,\gamma}$. 
In particular, many on-policy algorithms, including PPO~\citep{JS-FW-PD:2017}, collect trajectories
under the current policy and form minibatches by sampling uniformly
from a rollout buffer. 
To characterize the resulting state distribution, consider data collected under a fixed policy $\pi$, starting from $s_0\sim d_0$. 
{
In episodic tasks, a new episode starts from $d_0$ whenever the absorbing state is reached, whereas in continuing tasks the trajectory proceeds without restarting.}
As the amount of collected data grows, the empirical state frequencies  within the buffer converge to an undiscounted state distribution $d_\pi $.
In continuing tasks, $d_\pi $ is the stationary distribution of $P^\pi$.
In episodic tasks, $d_\pi (z)\triangleq 0$, while for $s\in\tilde{\mathcal S} $, $d_\pi (s)$  is obtained by normalizing the undiscounted cumulative occupancy measure $\sum_{t=0}^{\infty}\mathbb{E} _{s_0\sim d_0}\left[\mathbb{P}(s_t=s\,|\, s_0,\pi)\right]$ over  $\tilde{\mathcal S} $, as detailed in \eqref{eq:Episodicdpi}.
\textcolor{black}{Consequently, uniform sampling from the rollout buffer  approximates an expectation under $d_\pi $ rather than $d_{\pi,\gamma} $. Replacing $d_{\pi,\gamma} $ by $d_\pi $ in~\eqref{eq:epJ} gives the biased policy-gradient direction:}
\begin{equation} \label{eq:epbiasedJ}   {\color{black}
	\hat{\nabla}^{\textup{e}}_\theta J \triangleq \mathbb{E}_{s \sim 
		d_\pi }\left[\sum_{a\in\mathcal{A}} \nabla_\theta \pi(a  \, | \,  s)  Q^\pi(s, a)\right].}
		\end{equation}
		\textcolor{black}{To isolate the effect of the distribution mismatch from finite-sample
estimation noise, we study the idealized biased update
\begin{equation} \label{eq:biasedPG}
	\theta_{t+1} =\theta_{t}+\eta_t \hat{\nabla}^{\textup{e}}_\theta J (\theta_t),
\end{equation}
where $\hat{\nabla}^{\textup{e}}_\theta J (\theta_t)$ is the expected biased direction rather
than a finite-sample estimate.}

\section{Main results} \label{sec:analysis}
\textcolor{black}{In this section, we analyze biased policy-gradient methods under distribution mismatch. We first consider tabular parameterizations,
where global convergence results can be established via off-policy convergence theory, and then derive exact asymptotic
objective-error rates for tabular softmax policies. We subsequently
move beyond the tabular setting and quantify the distribution and
gradient mismatches under more general policy parameterizations.}
{\color{black} Throughout the analysis, we use the following standard assumptions for episodic and continuing MDPs, respectively. 
For episodic MDPs, we adopt the following  absorption condition, as in~\citep{SW-LS-GT:2022}.

\textcolor{black}{
	\begin{assumption}[Almost-sure absorption]\label{assumption:absorbing}
		For episodic MDPs, the absorbing state $z$ is reached almost surely from every initial state under any stationary policy $\pi$.
	\end{assumption}
}
For continuing MDPs,  we impose the following standard ergodicity condition, as in Che~et~al.~\citeyearpar{FC-MG-AM:2023} and Wu~et~al.~\citeyearpar{SW-LS-GT:2022}.
\begin{assumption}[Ergodicity]
	\label{assumption:ergodicity}
	For continuing MDPs, the Markov chain $P^\pi$ is irreducible and aperiodic under every stationary policy $\pi$.
\end{assumption}
}

{

\subsection{Tabular policy parameterization} \label{subsec:tabular}
We begin with two widely studied tabular policy parameterizations: the {direct} and {softmax parameterizations}. For each state-action pair $(s, a) \in \mathcal{S} \times \mathcal{A}$, the parameter $\theta_{s,a}$ 
{represents an action probability in the direct parameterization and a logit in the softmax parameterization.}  Both  formulations have been widely used to obtain theoretical insights into policy-gradient methods~\citep{JM-CX-DS:2020,AA-SK-GM:2021,GL-YW-YC:2021,GL:2022}.

\textbf{Direct parameterization:}
Under the direct parameterization,
$\pi_\theta(a\,|\, s)=\theta_{s,a}$.
The biased update direction in~\eqref{eq:epbiasedJ} has components
\begin{equation}\label{eq:gradientdirect}
    (\hat{\nabla}^{\textup{e}}_\theta J )_{s,a}=d_\pi(s)Q^\pi(s,a).
\end{equation}
To maintain valid action probabilities, each update is projected 
onto the policy simplex
$\Pi=\prod_{s\in\mathcal S}\Delta(\mathcal A)$:
\begin{equation}\label{eq:project}
    \theta_{t+1}
    =\operatorname{Proj} _\Pi\left(\theta_t+\eta\hat{\nabla}^{\textup{e}}_\theta J (\theta_t)\right),
\end{equation}
where $\operatorname{Proj} _\Pi$ is  Euclidean projection onto $\Pi$, and $\theta_0\in\Pi$.

\textbf{Tabular softmax parameterization:}
Under the tabular softmax parameterization, 
    $\pi_\theta(a\,|\, s)
    =\frac{e^{\theta_{s,a}}}
    {\sum_{a'\in\mathcal A}e^{\theta_{s,a'}}}$.
The corresponding biased update direction has components
\begin{equation}\label{eq:gradienttabular}
    (\hat{\nabla}^{\textup{e}}_\theta J )_{s,a}
    =d_\pi(s)\pi(a\,|\, s)A^\pi(s,a),
\end{equation}
where $A^\pi(s,a)=Q^\pi(s,a)-V^\pi(s)$ is the advantage
function. The parameters are updated according
to~\eqref{eq:biasedPG} without projection.
{Before studying asymptotic rates, we first show global objective-value convergence under biased updates for both parameterizations by verifying that the state distributions $d_{\pi_t}$ satisfy the coverage condition required in the work of Laroche and Tachet des Combes~\citep{RL-RT:2021}.}
\begin{theorem}[Global convergence under tabular parameterizations]
\label{thm:convergenceoftabular}
Suppose that \cref{assumption:absorbing} holds in episodic MDPs or 
\cref{assumption:ergodicity} holds in continuing
MDPs.  Let $\{\theta_t\}$ be generated by~\eqref{eq:project} under the direct
parameterization or by~\eqref{eq:biasedPG} under the softmax
parameterization, with any constant stepsize $\eta>0$. Then 
 $J(\pi_{\theta_t})$ converges
asymptotically to the optimal value $J^*$.
\end{theorem}
\begin{proof}
The proof is postponed to~\cref{prf:convergenceoftabular}. 
\end{proof}
\cref{thm:convergenceoftabular} ensures global
objective-value convergence, but does not quantify how the 
distribution mismatch affects the asymptotic convergence speed.
Existing analyses establish $O(1/t)$ convergence guarantees 
for tabular softmax policy gradient methods with exact
gradients~\citep{JM-CX-DS:2020,JL-WL-KW:2024}. For off-policy policy updates under general state weights, Laroche and Tachet des Combes~\citep{RL-RT:2021} further derive asymptotic error bounds. 
These results characterize the convergence
order but do not identify the exact leading-order coefficient of the
objective error. 
We next derive these coefficients for both the
normalized exact and biased updates, which in turn enables a direct
comparison of their asymptotic convergence speeds. 
Following~\citep[Assumption~A8]{RL-RT:2021}, we impose
the following unique-optimal-action assumption at 
the states in $\tilde{\mathcal S} $.
\begin{assumption}[Unique optimal action]
\label{assumption:unique_optimal_action}
For every $s\in\tilde{\mathcal S} $, there exists
a unique action $a^*(s)$ such that $Q^*(s,a^*(s))=V^*(s)$.
\end{assumption}
Under this assumption, we obtain the exact first-order asymptotics of
the objective errors.
\begin{theorem}[Exact asymptotic objective-error rates under tabular softmax parameterization]
\label{thm:softmax_asymptotic}
Suppose that \cref{assumption:unique_optimal_action}
holds, together with \cref{assumption:absorbing}
in episodic MDPs or \cref{assumption:ergodicity}
in continuing MDPs. Consider
the tabular softmax parameterization, and let $\pi^*$ denote an optimal policy. Let 
$\{\pi_t^{\mathrm e}\}$ and $\{\pi_t^{\mathrm b}\}$
be generated by the normalized exact  and biased policy-gradient updates, respectively, with the same
constant stepsize $\eta>0$. Then
\begin{equation}\label{eq:softmax_asymptotic_exact}
   \lim_{t\to\infty}
    t\bigl[J^*-J(\pi_t^{\mathrm e})\bigr]
    =
    \frac{\kappa_{\pi^*}(|\mathcal A|-1)|\tilde{\mathcal S} |}
    {\eta|\mathcal A|},
\end{equation}
and
\begin{equation}\label{eq:softmax_asymptotic_biased}
    \lim_{t\to\infty}
    t\bigl[J^*-J(\pi_t^{\mathrm b})\bigr]
    =
   \frac{\kappa_{\pi^*}(|\mathcal A|-1)}
    {\eta|\mathcal A|}
    \sum_{s\in\tilde{\mathcal S} }
    \frac{d_{\pi^*,\gamma}(s)}{d_{\pi^*}(s)}.
\end{equation}
Here, $\kappa_{\pi^*}$ is the normalization factor defined in
\eqref{eq:kappa}.
\end{theorem}
\begin{proof}
The proof is postponed to~\cref{prf:softmax_asymptotic}. 
\end{proof}

\cref{thm:softmax_asymptotic} gives the exact leading-order
coefficient $C$ in 
$J^*-J(\pi_t)=C/t+o(1/t)$. For the normalized exact and biased updates, the corresponding
coefficients are given by
\eqref{eq:softmax_asymptotic_exact} and
\eqref{eq:softmax_asymptotic_biased}, respectively.
Their ratio yields the following direct asymptotic comparison.

\begin{corollary}[Asymptotic comparison of biased
and normalized exact updates]
\label{cor:softmax_asymptotic_comparison}
Suppose that \cref{assumption:unique_optimal_action}
holds, together with \cref{assumption:absorbing}
in episodic MDPs and \cref{assumption:ergodicity}
in continuing MDPs. If $|\mathcal A|>1$,
then
\begin{equation}\label{eq:softmax_asymptotic_ratio}
    \lim_{t\to\infty}
    \frac{J^*-J(\pi_t^{\mathrm b})}
         {J^*-J(\pi_t^{\mathrm e})}
    =
    \frac{1}{|\tilde{\mathcal S} |}
    \sum_{s\in\tilde{\mathcal S} }
    \frac{d_{\pi^*,\gamma}(s)}{d_{\pi^*}(s)}.
\end{equation}
\end{corollary}
\begin{proof}
    Since $|\mathcal A|>1$, the limit in \eqref{eq:softmax_asymptotic_exact}  is strictly positive. Dividing the limit in \eqref{eq:softmax_asymptotic_biased} by that in \eqref{eq:softmax_asymptotic_exact} therefore yields \eqref{eq:softmax_asymptotic_ratio}.
\end{proof}
Consequently, under the same constant stepsize, the biased
update is asymptotically faster than the normalized
exact update when   $\sum_{s\in\tilde{\mathcal S} }
    \frac{d_{\pi^*,\gamma}(s)}{d_{\pi^*}(s)}
    <|\tilde{\mathcal S} |$, and asymptotically slower when the inequality is reversed.

\begin{remark}[General parameterizations]
Even under restricted, non-tabular parameterizations, biased updates may  attain global optimality within the policy class and may converge faster than normalized
exact  updates in some problems; see \cref{example:biased} for an example.
\end{remark}
}

\subsection{Bounds for distribution mismatch} \label{subsec:bounds}

We now move beyond the tabular setting to general policy parameterizations, for which the preceding global
convergence and exact asymptotic rate
characterizations need not hold. As a first
step, we quantify the underlying state distribution mismatch between
the normalized discounted distribution $d_{\pi,\gamma}$ and the
undiscounted distribution $d_\pi$. These bounds will serve as the
basis for analyzing the induced policy gradient mismatch in the next
subsection.
\textcolor{black}{ Wu et al.~\citeyearpar{SW-LS-GT:2022} also quantify errors  arising from the distribution mismatch. However, in episodic MDPs, their analysis concerns unnormalized discounted and undiscounted state occupancy measures. We instead compare the normalized distributions $d_{\pi,\gamma}$ and $d_\pi$, which are the distributions directly
associated with the normalized exact and biased policy gradient
directions considered in this paper.  In particular, $d_\pi$
describes the limiting state frequencies induced by uniform sampling
from the rollout buffer. 
For continuing MDPs, we directly apply the
distribution mismatch bound of Wu et al.~\citeyearpar[Theorem~4.4]{SW-LS-GT:2022}.}

\textbf{Episodic MDPs}: 
We first bound the distribution mismatch for episodic MDPs under \cref{assumption:absorbing}.
\begin{theorem}[Mismatch bound for episodic MDPs] \label{thm:boundforepisodic} For episodic MDPs under \cref{assumption:absorbing}, there exist constants $m>0$ and $\alpha\in(0,1)$, independent of $\pi$ and $\gamma$, such that for every stationary policy $\pi$ and every $\gamma\in[0,1)$,
\begin{equation*}
	\|d_{\pi,\gamma} -d_\pi \|_1 \le 		\frac{2 m}{1-\alpha}(1-\gamma).
\end{equation*}
\end{theorem}
\begin{proof}
The proof is postponed to~\cref{prf:boundforepisodic}. 
\end{proof}

\textbf{Continuing MDPs}: We next consider continuing MDPs, where  the undiscounted distribution $d_\pi $ coincides with the stationary distribution of  the transition matrix $P^\pi$. 
\textcolor{black}{We apply~\citep[Theorem~4.4]{SW-LS-GT:2022}.
As  \cref{lem:uniform_geometric_convergence}  in \cref{appendix:auxiliary} shows, under \cref{assumption:ergodicity}, the geometric ergodicity condition required therein holds with constants that are uniform over all
stationary policies. We therefore obtain the following result. }
\begin{theorem}[Mismatch bound for continuing MDPs~{\citep[Theorem~4.4]{SW-LS-GT:2022}}]\label{thm:boundforcontinuing}  For continuing MDPs under \cref{assumption:ergodicity}, there exist constants $D>0$ and $\beta\in(0,1)$, independent of $\pi$ and $\gamma$,   such that for every stationary policy $\pi$ and every $\gamma\in[0,1)$,
\begin{equation*}
	\| d_{\pi,\gamma} - d_\pi \|_{1} \leq  \frac{2D}{1-\beta \gamma}(1-\gamma).
\end{equation*}
\end{theorem}
Taken together, \cref{thm:boundforepisodic,thm:boundforcontinuing} show that in both
episodic and continuing MDPs, the mismatch between discounted and undiscounted state distributions vanishes as $\gamma \to 1$.

\subsection{Gradient mismatch induced by distribution mismatch}\label{subsec:biasedgradient}
Building on the distribution mismatch bounds in \cref{subsec:bounds}, we now quantify the discrepancy between the normalized exact policy-gradient direction \(\nabla_\theta^{\textup{e}} J \) and the biased  direction \(\hat{\nabla}^{\textup{e}}_\theta J \). 
We impose the following boundedness condition on the policy derivatives. \begin{assumption}[Bounded policy derivatives] \label{assumption:parameterization} There exists a constant $G>0$ such that, for every $(s,a)\in\mathcal S\times\mathcal A$ and every $\theta$,  $\nabla_\theta\pi(a \, | \, s)$ exists and satisfies $\|\nabla_\theta\pi(a  \, | \,  s)\|_2\leq G$.
\end{assumption}

\textcolor{black}{\cref{assumption:parameterization} is weaker than commonly used bounded-score condition requiring  $\left\|\nabla_\theta \log \pi(a \,|\, s)\right\|_2$ to be uniformly bounded~\citep{MP-DB-MR:2018,KZ-AK-TB:2020,SM-DK:2024}.}
Under this assumption, we first establish the gradient-mismatch bound
for episodic MDPs.
\begin{theorem}[Episodic gradient mismatch bound]
\label{thm:gradbias_ep}
For episodic MDPs under \cref{assumption:absorbing,assumption:parameterization}, let $m>0$ and $\alpha\in(0,1)$ be the constants in \cref{thm:boundforepisodic}. Then, for every $\theta$  with $\pi=\pi_\theta$,
\[
\|\nabla_\theta^{\textup{e}} J -\hat{\nabla}^{\textup{e}}_\theta J \|_2
\le \varepsilon_{\mathrm{e}}(\gamma),
\]
where $\varepsilon_{\mathrm{e}}(\gamma)= \frac{2G|\mathcal A|\,m^2R_{\max}}{(1-\alpha)^2}
(1-\gamma)$.%
\end{theorem}
\begin{proof}
The proof is postponed to~\cref{proof:gradbias_ep}. 
\end{proof}
For continuing MDPs, we obtain the following counterpart.
\begin{theorem}[Continuing gradient mismatch bound]
\label{thm:gradbias_cont}
For continuing MDPs under \cref{assumption:ergodicity,assumption:parameterization}, let $D>0$ and $\beta\in(0,1)$ be the constants in \cref{thm:boundforcontinuing}. Then, for every $\theta$ with $\pi=\pi_\theta$,
\[
\left \| \nabla_\theta^{\textup{e}} J - \hat{\nabla}^{\textup{e}}_\theta J \right\|_2  \leq \varepsilon_{\mathrm{c}}(\gamma),
\]
where $\varepsilon_{\mathrm{c}}(\gamma)=  G|\mathcal{A}|\left(2 R_{\max }+\frac{8\gamma D R_{\max }}{1-\beta\gamma}\right)\frac{2 D}{1-\beta\gamma}(1-\gamma)
$.%
\end{theorem}
\begin{proof}
The proof is postponed to~\cref{proof:gradbias_cont}. 
\end{proof}\cref{thm:gradbias_ep,thm:gradbias_cont} further show that the gradient mismatch is $O(1-\gamma)$ in both episodic
and continuing MDPs, uniformly over $\theta$. We next examine what this gradient-level
control implies for the optimization behavior of the biased iterates.

\subsection{Optimization implications of gradient mismatch} \label{subsection:performance}
To translate the gradient-mismatch bounds into optimization
guarantees for the biased iterates, we additionally assume that the objective has a Lipschitz-continuous gradient.
\begin{assumption}[$L$-smoothness]
\label{assumption:smoothness}
There exists a constant $L>0$ such that, for all $\theta_1$ and $\theta_2$,
$$\left\|\nabla J(\theta_1)-\nabla J(\theta_2)\right\|_2
\le
L\left\|\theta_1-\theta_2\right\|_2.$$
\end{assumption}

\textcolor{black}{\cref{assumption:smoothness} is a standard smoothness condition in policy gradient convergence analyses~\citep{JM-CX-DS:2020,AA-SK-GM:2021,KZ-AK-TB:2020}.
Together with \cref{assumption:parameterization} and the preceding gradient mismatch bounds, it yields the following approximate
stationarity guarantee.}
{\color{black}
\begin{theorem}[Approximate guarantee for biased policy gradient iterates]
	\label{thm:performance}
	Suppose that \cref{assumption:parameterization,assumption:smoothness} hold, together with \cref{assumption:absorbing} in episodic MDPs or \cref{assumption:ergodicity} in continuing MDPs.
	Consider the iterates generated by~\eqref{eq:biasedPG} with stepsizes satisfying  $0<\eta_t\leq1/L$ for all $t$ and $\sum_{t=0}^{\infty}\eta_t=\infty$.
	Then
	\[
	\liminf_{t\to\infty}\|\nabla_\theta^{\textup{e}} J (\theta_t)\|_2^2
	\leq\varepsilon_\gamma^2=O\bigl((1-\gamma)^2\bigr),
	\]
	where $\varepsilon_\gamma=\varepsilon_{\mathrm e}(\gamma)$ in episodic MDPs and $\varepsilon_\gamma=\varepsilon_{\mathrm c}(\gamma)$ in continuing MDPs, with $\varepsilon_{\mathrm e}(\gamma)$ and
$\varepsilon_{\mathrm c}(\gamma)$ given in \cref{thm:gradbias_ep,thm:gradbias_cont}, respectively.
\end{theorem}

\begin{proof}
	The proof is postponed to~\cref{appendix:performance}.
\end{proof}

}

\cref{thm:performance} shows that biased iterates retain an approximate stationarity guarantee with respect to the normalized exact policy-gradient direction. In particular, the guaranteed
residual in the squared norm is $O((1-\gamma)^2)$ in both episodic
and continuing MDPs and therefore vanishes as $\gamma\to1$.
\section{Numerical Examples} \label{sec:simulation}
In this section, we present numerical experiments to illustrate the main theoretical results.
{ We first consider tabular parameterizations, where the update directions can be evaluated exactly without sampling
noise. We examine global objective-value convergence
and then verify
the exact asymptotic coefficients derived for tabular softmax policies.}
We subsequently consider neural-network-based actor-critic methods in
both continuing and episodic environments to investigate the empirical
effect of distribution mismatch under general function approximation. 

\subsection{Tabular parameterizations}\label{subsection:tabular}

Since the tabular examples have finite state and action spaces of moderate size, both the biased and normalized exact policy-gradient directions are computed directly from their respective analytical expressions rather than estimated from samples. This eliminates finite-sample noise and allows a direct comparison of the two update
directions.

\begin{figure*}[!ht]
\centering
\includegraphics[width=0.92\linewidth]{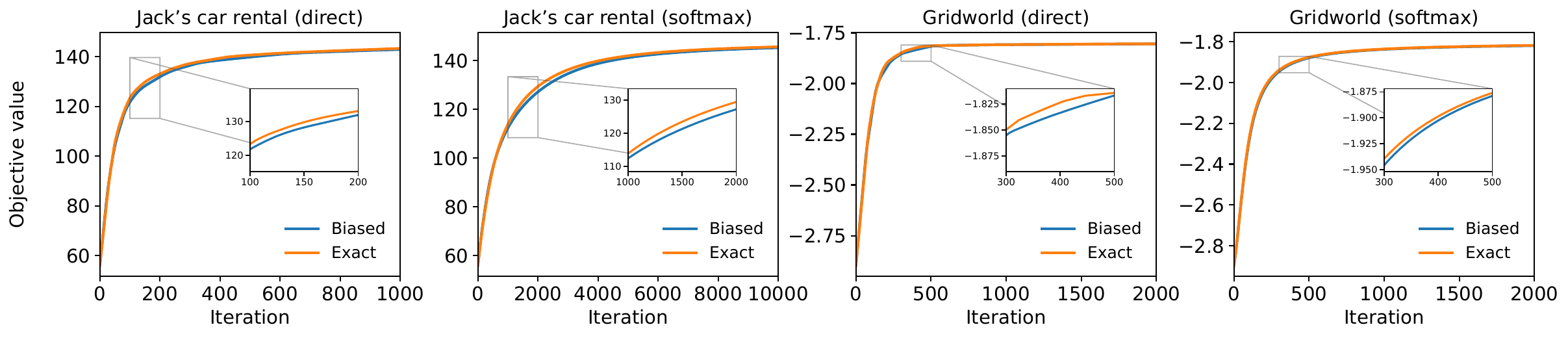}
\caption{
Objective-value convergence under biased and normalized exact
policy-gradient updates with $\gamma=0.7$.
}
\label{figure:tabularConvergence}
\end{figure*}

\textbf{Global convergence}:
We first consider two standard benchmarks to illustrate the global convergence of biased updates under the direct and tabular softmax parameterizations. For continuing MDPs, we adopt Jack's car rental problem \citep[Example 4.2]{RS-AB:2018}, where each location can hold at most $5$ cars and at most $3$ cars can be transferred per day. For episodic MDPs, we consider a stochastic $4\times4$ gridworld based on \citep[Example 4.1]{RS-AB:2018}, with two terminal states at opposite corners. Each transition incurs a reward of $-1$, and stochastic perturbations introduce additional rightward and downward movements. The results for both parameterizations are shown in \cref{figure:tabularConvergence}.

\begin{figure*}[t]
\centering    \includegraphics[width=0.92\linewidth]{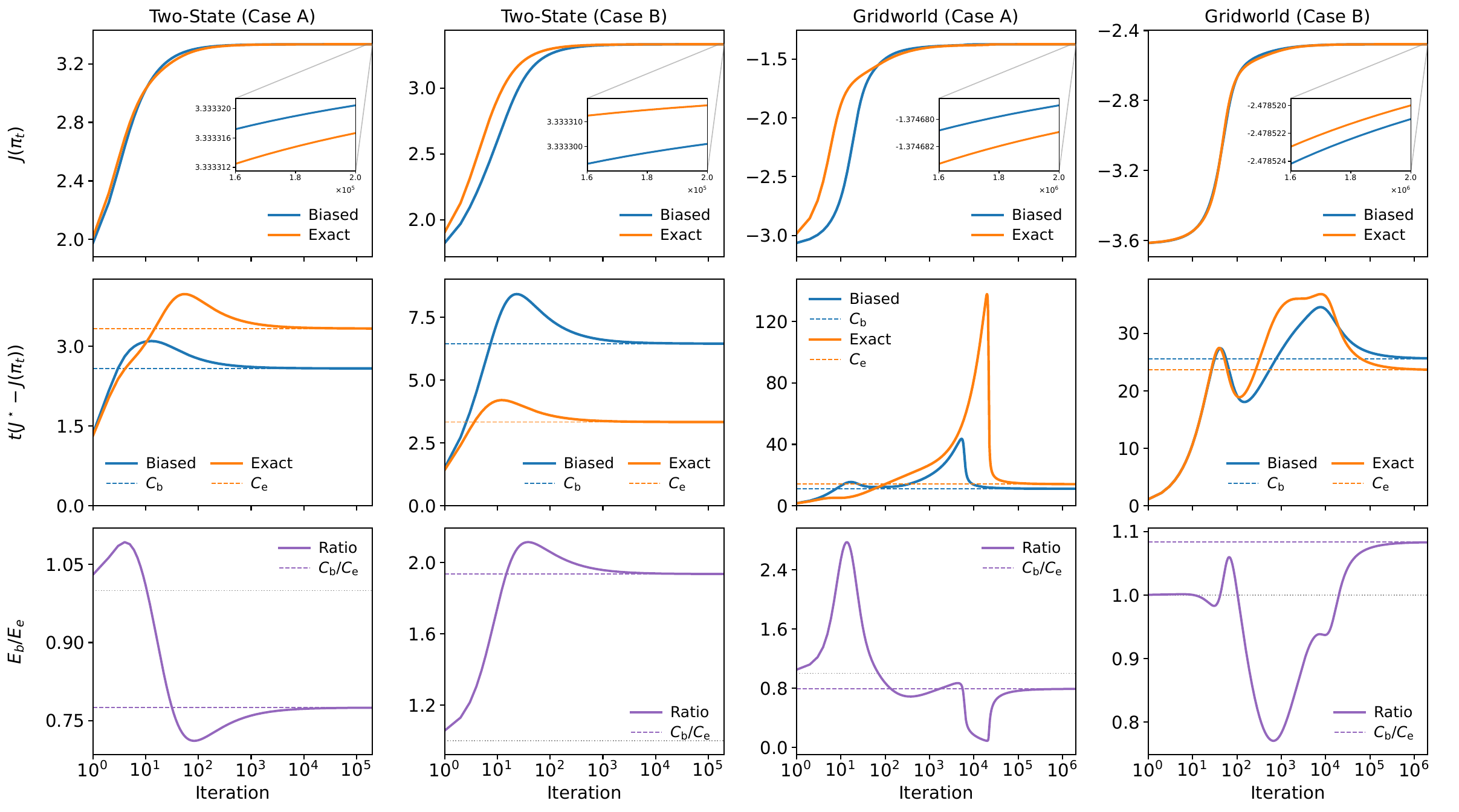}

\caption{Asymptotic convergence under the tabular softmax parameterization. Top: objective values with zoomed-in views. Middle: scaled objective errors and their theoretical limits. Bottom: objective-error ratios and their theoretical limits.}
\label{figure:asymptoticComparison}
\end{figure*}

\textbf{Exact asymptotic comparison}:
We next construct two illustrative MDPs, a two-state continuing MDP and a stochastic episodic gridworld, to examine the exact asymptotic convergence behavior characterized in \cref{thm:softmax_asymptotic,cor:softmax_asymptotic_comparison}. Detailed configurations are provided in
Appendix~\ref{app:experimental_configurations}.
Let $E_t^m:=J^*-J(\pi_t^m)$ for $m\in\{\mathrm e,\mathrm b\}$ denote the objective errors of the normalized exact and biased updates, respectively.
Let $C_{\mathrm e}$ and $C_{\mathrm b}$ be the corresponding theoretical asymptotic 
coefficients given by
\eqref{eq:softmax_asymptotic_exact} and
\eqref{eq:softmax_asymptotic_biased}, respectively.
For each MDP, we construct two cases by varying the initial state distribution, such that $C_{\mathrm b}/C_{\mathrm e}<1$ in Case A and $C_{\mathrm b}/C_{\mathrm e}>1$ in Case B. 
We examine whether the scaled objective errors $tE_t^m$ converge to the coefficients $C_m$ and whether the objective-error ratio $E_t^{\mathrm b}/E_t^{\mathrm e}$ converges to with the theoretical limit $C_{\mathrm b}/C_{\mathrm e}$ in \eqref{eq:softmax_asymptotic_ratio}. The results with are shown in \cref{figure:asymptoticComparison}.

\subsection{General policy parameterizations}
To complement the tabular experiments,  we also consider a neural-network-based actor-critic implementation  in both continuing and episodic MDPs. %
Trajectories are first collected into a rollout buffer, from which mini-batches are sampled uniformly to update the actor. This corresponds to the biased sampling scheme. As a corrected baseline, we reweight each transition by its temporal
discount factor $\gamma^t$, following the correction principle of Zhang et al.~\citeyearpar{SZ-RL-VH:2022}. To remain consistent with the theoretical formulation, the resulting weights are further normalized to provide a sample-based
estimator of the normalized exact policy-gradient direction\(\nabla_\theta^{\textup{e}} J \). 

For all tasks, the actor and critic are represented by separate two-hidden-layer multilayer perceptrons with hyperbolic-tangent activations and orthogonal initialization. We use stochastic gradient descent, one optimization epoch per rollout, generalized advantage estimation with \(\lambda = 0.95\), a value-loss coefficient of $0.5$, no entropy regularization, and no gradient clipping. {The curves report the mean return across $10$ random seeds, with shaded regions indicating one standard deviation.} Returns are discounted and the curves are shown without temporal smoothing. {Task-specific network and optimization hyperparameters are provided in the task descriptions.}

\begin{figure*}[t]
\centering

\includegraphics[width=0.92\linewidth]{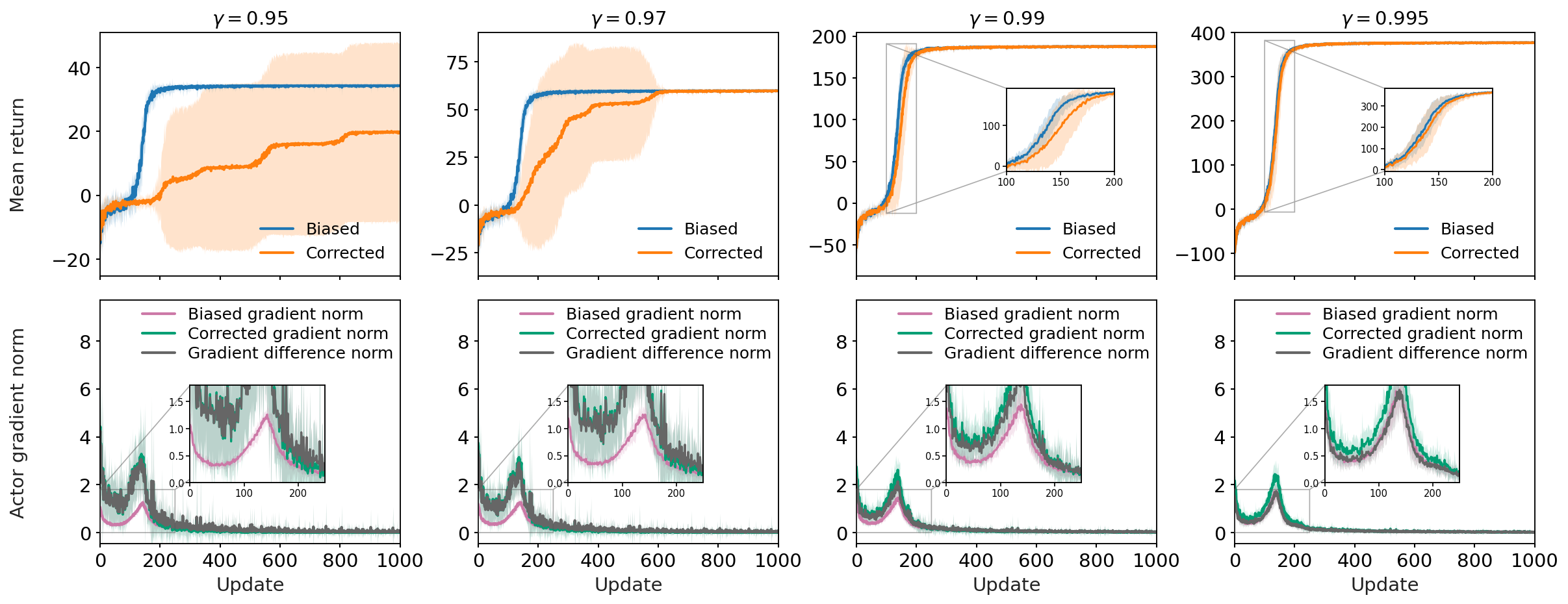}

\caption{Actor-critic results on \textbf{CliffWalking} with a neural network parameterization.
	Top: mean discounted returns under biased and corrected training.
	Bottom: $\ell_2$-norms of the biased and corrected actor gradients and of their difference, all evaluated along the biased-training trajectory.}
\label{figure:cliffwalking}
\end{figure*}

\textbf{Continuing MDPs}: We first consider a continuing $4\times4$ \textbf{CliffWalking} environment.  An ordinary step, a transition into the cliff, and a transition to the goal yield rewards $-0.1$, $-5$, and $10$, respectively.  Both cliff and goal transitions return the agent to the start rather than terminating the process. Rollouts are truncated after $1000$ steps for data collection.  States are one-hot encoded. { Both the actor and critic use $64$ units per hidden layer. We use a learning rate of $0.005$, a rollout size of $2048$, and a batch size of $256$.}
Results are shown in \cref{figure:cliffwalking}.

\begin{figure*}[t]
\centering

\includegraphics[width=0.92\linewidth]{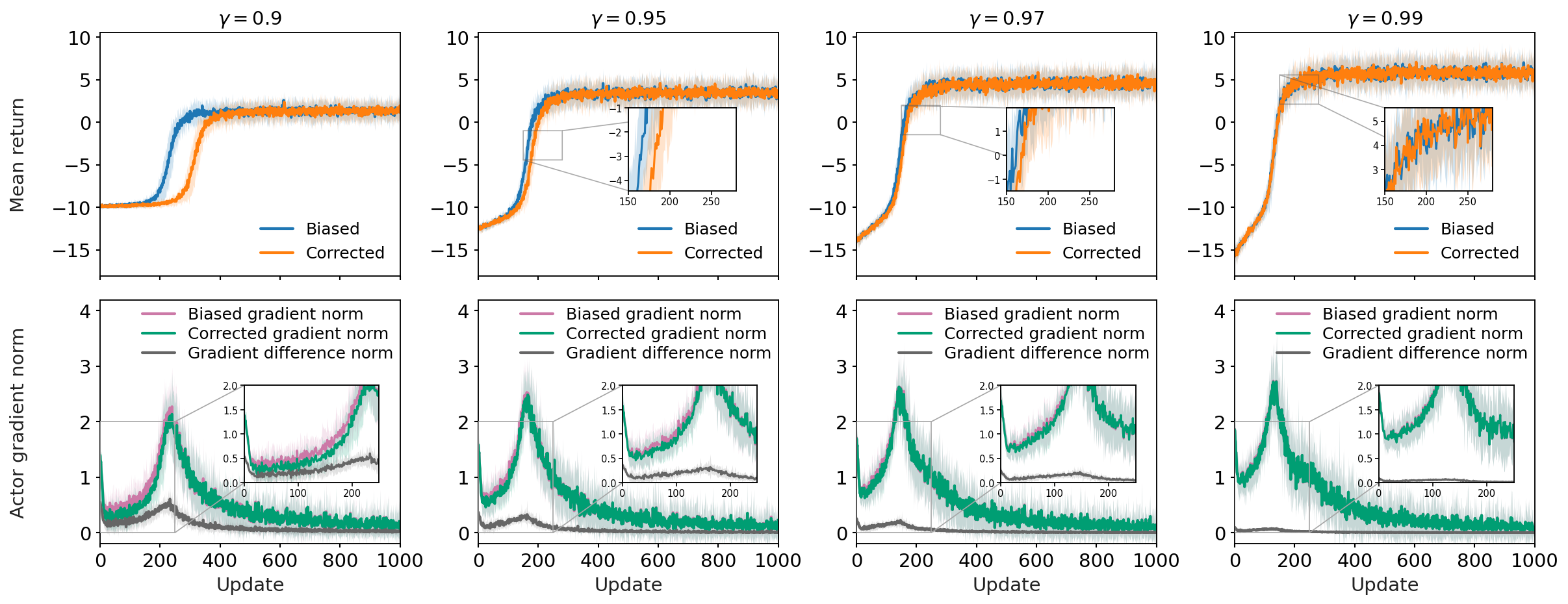}

\caption{Actor-critic results on \textbf{FrozenLake} with a neural network parameterization.
	Top: mean discounted returns under biased and corrected training.
	Bottom: $\ell_2$-norms of the biased and corrected actor gradients and of their difference, all evaluated along the biased-training trajectory.}
\label{figure:frozenlake}
\end{figure*}

\textbf{Episodic MDPs}: We next consider a stochastic $4\times4$ \textbf{FrozenLake} environment with intended-transition probability $0.8$.  Reaching the goal, entering a hole, and taking an ordinary step yield rewards $20$, $-10$, and $-1$, respectively. Reaching either the goal or a hole terminates the episode.  States are one-hot encoded. { Both the actor and critic use $64$ units per hidden layer. We use a learning rate of $0.005$, a rollout size of $512$, and a batch size of $128$.}
The corresponding results are shown in \cref{figure:frozenlake}.

\begin{figure*}[t]
\centering    \includegraphics[width=0.92\linewidth]{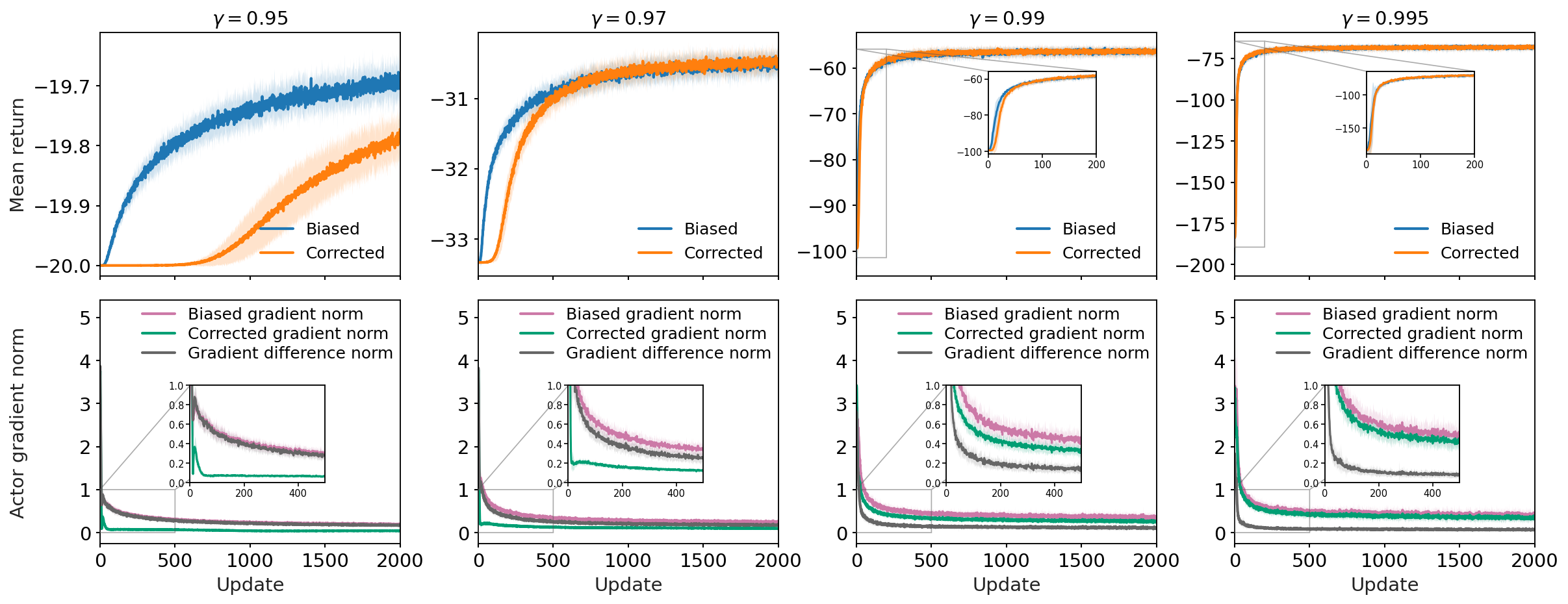}

\caption{Actor-critic results on \textbf{Acrobot-v1} with a neural network parameterization.
	Top: mean discounted returns under biased and corrected training.
	Bottom: $\ell_2$-norms of the biased and corrected actor gradients and of their difference, all evaluated along the biased-training trajectory.}
\label{figure:acrobot}
\end{figure*}

Finally, We consider the standard Gymnasium \textbf{Acrobot-v1} benchmark, which has a continuous state space and terminates when the end effector reaches the target height. Continuous observations are provided directly to the networks. { Both the actor and critic use $512$ units per hidden layer. We use a learning rate of $0.001$, a rollout size of $8192$, and a batch size of $512$.} 
Results are shown in \cref{figure:acrobot}.

{\color{black}
\subsection{Results and discussion}
The numerical experiments give the following observations:
\begin{itemize}
	\item \textbf{Global convergence under tabular parameterizations.}
	As shown in \cref{figure:tabularConvergence},  biased and normalized
    exact updates  approach the same optimal objective value in both episodic and continuing tasks  under the direct and tabular softmax parameterizations, although their optimization trajectories differ.
	For the biased updates, this behavior is consistent with \cref{thm:convergenceoftabular}.

     \item \textbf{Exact asymptotic rates and comparison.}
    In \cref{figure:asymptoticComparison}, the scaled objective
    errors $tE_t^{\mathrm e}$ and $tE_t^{\mathrm b}$ approach
    their respective theoretical coefficients $C_{\mathrm e}$
    and $C_{\mathrm b}$ in 
    \cref{thm:softmax_asymptotic}. The objective-error ratios
    also approach the limits predicted by
    \cref{cor:softmax_asymptotic_comparison}. These observations hold in both continuing and episodic MDPs. Moreover, the objective curves show faster asymptotic decay for the biased updates in Case A, where $C_{\mathrm b}/C_{\mathrm e}<1$, and slower asymptotic decay in Case B, where $C_{\mathrm b}/C_{\mathrm e}>1$. 
    Together, these results support our characterization of when distribution mismatch accelerates objective-value convergence under tabular softmax parameterization with the same constant stepsize.
    
	\item \textbf{Reduced mismatch at larger discount factors.}
	In \cref{figure:cliffwalking,figure:frozenlake,figure:acrobot}, the biased and corrected training trajectories generally become more similar as $\gamma$ approaches one. Meanwhile, the norm of the difference between the corresponding actor gradient estimates, evaluated along the same biased-training trajectory, generally decreases with increasing $\gamma$. This trend is consistent with the gradient mismatch bounds in \cref{thm:gradbias_ep,thm:gradbias_cont}. The Acrobot experiment in \cref{figure:acrobot} further illustrates that a similar empirical trend can arise with continuous state spaces beyond the finite-state setting of our theoretical analysis.
	\item \textbf{Potential advantages under neural-network parameterizations.}
	As shown in \cref{figure:cliffwalking,figure:frozenlake,figure:acrobot}, the biased updates can learn faster than their corrected counterparts at smaller tested discount factors. This complements the example in Appendix~\ref{example:biased}, which illustrates that distribution mismatch may accelerate convergence. However, the advantage is not universal: \cref{figure:cliffwalking} shows faster initial improvement under corrected updates in some settings.
\end{itemize}
}

\section{Conclusion}\label{sec:conclusion}
We study the impact of  state distribution mismatch on policy gradient methods for discounted MDPs.  Under direct and tabular softmax parameterizations, we establish objective-value convergence to the global optimum despite biased updates. For tabular softmax policies, we derive the exact leading-order coefficients of
the asymptotic objective errors for biased and normalized exact
updates. Their ratio provides an explicit comparison of the
asymptotic convergence speeds of the two updates under the same
constant stepsize. 
For general parameterizations, we derive explicit bounds on the
mismatch between discounted and undiscounted state distributions and
on the resulting normalized policy-gradient mismatch in both
episodic and continuing MDPs.  Under an additional smoothness assumption, we further establish an approximate-stationarity guarantee for the biased iterates, with an error bound that vanishes as the discount factor approaches one in both settings.
Numerical experiments illustrate
the theoretical results and the effects of distribution mismatch
under both tabular and neural-network policy parameterizations.

\appendix
\crefalias{section}{appendix}
\crefalias{subsection}{appendix}
\crefname{appendix}{Appendix}{Appendices}
\Crefname{appendix}{Appendix}{Appendices}
\begin{table}[ht]
\caption{Notation}\label{tab:notation}
\centering
\begin{tabularx}{\columnwidth}{@{}lX@{}}
	\hline
	\textbf{Notation} & \textbf{Meaning} \\
	\hline
	$J(\pi)$ & Objective function for policy $\pi$. \\
	$J(\theta)$ & Shorthand for $J(\pi_{\theta})$.\\
	$P^\pi$ & Transition matrix corresponding to policy $\pi$.
	\\
    
    $\tilde{\mathcal S} $ &
Relevant state set: $\mathcal S$ in continuing MDPs
and $\mathcal S\setminus\{z\}$ in episodic MDPs. \\

	$\tilde{P}^\pi$ & Principal submatrix of $P^\pi$ indexed by $\tilde{\mathcal S} $ in episodic MDPs.  It is a row-substochastic matrix.\\ 
	$d_0$ & Initial state distribution. $d_0(z)=0$ in episodic MDPs; $d_0(s)>0$ for every $s\in\tilde{\mathcal S} $.  \\
	$\tilde{d}_0$ & Restriction of $d_0$ to $\tilde{\mathcal S} $ in episodic MDPs, satisfying $\tilde{d}_0 ^\top\text{\usefont{U}{bbold}{m}{n}1}=1$. 
	\\
	$\mu_{\pi,\gamma}$ & Unnormalized discounted state occupancy measure defined in \eqref{eq:mu}. $\mu_{\pi,\gamma}$ is not a valid probability distribution.   \\

    $\kappa_\pi$ & Normalization factor $\sum_{s\in\tilde{\mathcal S} }\mu_{\pi,\gamma}(s)$, defined in \eqref{eq:kappa}. \\

	$d_{\pi,\gamma} $ & Discounted state distribution used in \eqref{eq:epJ}, with $d_{\pi,\gamma} (s)=\mu_{\pi,\gamma}(s)/\kappa_\pi$ for $s\in\tilde{\mathcal S} $ and $d_{\pi,\gamma} (z)=0$ in episodic MDPs. \\
	
	$\tilde{d}_{\pi,\gamma} $ & Restriction of $d_{\pi,\gamma} $ to $\tilde{\mathcal S} $ in episodic MDPs.  \\
	
	$d_\pi $ & Undiscounted state distribution. In continuing tasks, $d_\pi $ is the stationary distribution of $P^\pi$. 
    In episodic MDPs, it is obtained by normalizing the undiscounted state occupancy \(\sum_{t=0}^\infty \mathbb{E} _{s_0 \sim d_0} \left[P^\pi(s_t = s \, | \, s_0)\right]\) over $\tilde{\mathcal S} $, with $d_\pi (z)=0$.
    {It is approximated empirically by uniformly sampling states from collected on-policy trajectories.}\\
	
	$\tilde{d}_\pi $ & Restriction of $d_\pi $ to $\tilde{\mathcal S} $ in episodic MDPs, given in \eqref{eq:Episodicdpi}.  \\
	$\nabla_\theta J $ & Objective gradient of $J(\theta)$, given in \eqref{eq:gradientJ}. \\
	
	$\nabla_\theta^{\textup{e}} J $ &  \textcolor{black}{ Normalized exact policy-gradient direction underlying many popular algorithms, defined in \eqref{eq:epJ}.}\\
	
	$\hat{\nabla}^{\textup{e}}_\theta J $  & Widely used biased version of $\nabla_\theta^{\textup{e}} J $, which replaces the inaccessible distribution $d_{\pi,\gamma} $ with $d_\pi $ in \eqref{eq:epJ}.\\
	
	\hline
\end{tabularx}
\end{table}

\section{Auxiliary lemmas}\label{appendix:auxiliary}
The notation used throughout the paper, including the appendices, is summarized in \cref{tab:notation}.
We first introduce several auxiliary results that will be used  in the subsequent proofs. For finite-state and finite-action episodic MDPs, the following  uniform finite-step absorption condition is known to be equivalent to \cref{assumption:absorbing}~\citep[Exercise~2.3]{DPB:2007}, \citep[Assumption~4.2.1]{DPB:2019}.
\begin{lemma}[Uniform absorption bound]
\label{lemma:absorbing}
For episodic MDPs under \cref{assumption:absorbing},  there exists an integer $m > 0$ and a positive real number $\alpha < 1$ such that $\mathbb{P}\left(s_m \neq z  \, | \,  s_0, \pi\right) \leq \alpha$ for all $s_0 \in \mathcal{S}$ and any stationary policy $\pi$.
\end{lemma}

We next present a geometric convergence bound for continuing MDPs, which follows from the standard uniform ergodicity result for finite-state and finite-action Markov decision processes~\citep[Theorem~3.4]{QH-WY:08}.

\begin{lemma}[Uniform geometric convergence]
\label{lem:uniform_geometric_convergence}
For continuing MDPs under Assumption~\ref{assumption:ergodicity}, there exist constants $D>0$ and $\beta\in(0,1)$, independent of the stationary policy $\pi$, such that
\begin{equation}\label{eq:geo}
	\left\|\nu^\top(P^\pi)^t-d_\pi^\top\right\|_{\mathrm{TV}}
	\le D\beta^t
\end{equation}
for every stationary policy $\pi\in\Pi$,  every probability distribution $\nu$ on $\mathcal S$, and  every $t\ge0$, where $\|\cdot\|_{\mathrm{TV}}$ is the total variation distance.
\end{lemma}

The following lemma bounds the expected absorption time for episodic MDPs.

\begin{lemma}[Upper bound on the expected absorption time]
\label{lem:absorbing_time}
For episodic MDPs under Assumption~\ref{assumption:absorbing}, let $\tau$ denote the hitting time of the absorbing state $z$.
Then
\begin{equation}\label{eq:expectationtau}
	\mathbb E_\pi[\tau  \, | \,  s_0]
	\le \frac{m}{1-\alpha},
	\qquad
	\forall s_0\in\mathcal S,\quad \forall\pi\in\Pi.
\end{equation}
\end{lemma}

\begin{proof}
By \cref{lemma:absorbing}, the non-absorption probability satisfies $\mathbb P(\tau>km\,|\, s_0,\pi)\leq\alpha^k$~\citep[Eq.~(4.5)]{DPB:2019}. Hence, using the standard tail-sum formula,
\[
\mathbb E_\pi[\tau\,|\, s_0]
=\sum_{t=0}^\infty\mathbb P(\tau>t\,|\, s_0,\pi)
\leq
m\sum_{k=0}^\infty\alpha^k
=\frac{m}{1-\alpha}.
\]
\end{proof}

Then we recall the performance difference lemma.
\begin{lemma}[Performance difference lemma~{\citep[Lemma 6.1]{SK-JL:2002}}]\label{lemma:performance_difference} Let $A^\pi(s,a)$ denote the advantage function associated with the state-action pair $(s,a)$. For any two policies $\pi$ and $\tilde{\pi}$,
\begin{equation*}\label{eq:difference1}
	J({\pi})-J(\tilde{\pi})= \sum_s \mu_{\pi,\gamma}(s) \sum_a \pi(a  \, | \,  s) \left[A^{\tilde{\pi}}(s, a)\right],   
\end{equation*}
where $\mu_{\pi,\gamma}$ is defined in \eqref{eq:mu}.
\end{lemma}
In episodic MDPs, \(A^{\tilde\pi}(z,a)=0\) for every policy \(\tilde\pi\) and action \(a\), so the absorbing state contributes zero to the performance difference. Since \(\tilde{\mathcal S} =\mathcal S\) in continuing MDPs, the difference can therefore be written in both settings as
\begin{equation}\label{eq:difference}
    J(\pi)-J(\tilde{\pi})=\sum_{s \in \tilde{\mathcal S} } \mu_{\pi, \gamma}(s) \sum_{a \in \mathcal{A}} \pi(a \, | \,  s) A^{\tilde{\pi}}(s, a) .
\end{equation}

\section{\texorpdfstring{Proof of \cref{subsec:tabular}}{Proof: subsec tabular}} \label{appendix:tabular}

{

\subsection{\texorpdfstring{Proof of \cref{thm:convergenceoftabular}}{Proof: thm domination}} \label{prf:convergenceoftabular}
The convergence result follows from the tabular off-policy policy-update
framework of~\citep{RL-RT:2021} once its cumulative state-coverage
condition is verified for the state weights induced by our sampling
scheme in our case. We therefore first establish uniform positive lower bounds for
the undiscounted state distribution $d_\pi$ under
\cref{assumption:absorbing,assumption:ergodicity}, respectively.
These bounds hold uniformly over all stationary policies and all $s\in \tilde{\mathcal S} $.

\begin{lemma}[Uniform positivity of state probabilities in episodic MDPs]
\label{cor:episodic_uniform_bound}
For episodic MDPs, suppose that  \cref{assumption:absorbing} holds. Then the undiscounted state distribution $d_\pi $ depends continuously on $\pi$ and satisfies
\[
\inf_{\pi\in\Pi}\min_{s\in\tilde{\mathcal S} }d_\pi(s)
\ge
\frac{1-\alpha}{m}\min_{s\in \tilde{\mathcal S} }d_0(s)>0,
\]
where $m$ and $\alpha$ are given in \cref{lemma:absorbing}.
\end{lemma}
\begin{proof}
For a stationary policy $\pi$, let $\tilde{P}^\pi $ denote the principal submatrix of $P^\pi$ indexed by $\tilde{\mathcal S} $. Since $z$ is absorbing,  $\tilde{P}^\pi $ is row-substochastic. Let $\tilde d_0$ and $\tilde d_\pi$ denote the restrictions of $d_0$ and $d_\pi$ to $\tilde{\mathcal S} $, respectively. 
Since $d_0(z)=d_\pi (z)=0$, both $\tilde d_0$ and $\tilde d_\pi$ are probability distributions. By~\cref{lemma:absorbing}, for  every stationary  policy $\pi$, $(\tilde{P}^\pi )^m \text{\usefont{U}{bbold}{m}{n}1}\leq \alpha \text{\usefont{U}{bbold}{m}{n}1}$. Hence, $\rho((\tilde{P}^\pi )^m) \leq \|(\tilde{P}^\pi )^m\|_\infty \leq \alpha<1$, and therefore $\rho(\tilde{P}^\pi )^m = \rho((\tilde{P}^\pi )^m) <1$.  Thus $\rho(\tilde{P}^\pi ) <1$, so $I-\tilde{P}^\pi $ is invertible and $(I-\tilde{P}^\pi )^{-1} =
\sum_{t=0}^\infty(\tilde{P}^\pi )^t$. %

By the definition of the undiscounted state distribution in episodic MDPs,
\begin{equation} \label{eq:Episodicdpi}
	\tilde{d}_\pi ^\top = \frac{\sum_{t=0}^\infty  \tilde{d}_0 ^\top (\tilde{P}^{\pi})^t }{\sum_{t=0}^\infty \tilde{d}_0 ^\top (\tilde{P}^{\pi})^t \text{\usefont{U}{bbold}{m}{n}1} } = \frac{\tilde{d}_0 ^\top(I-\tilde{P}^\pi )^{-1}}{\tilde{d}_0 ^\top(I-\tilde{P}^\pi )^{-1}\text{\usefont{U}{bbold}{m}{n}1}}.  
\end{equation}
Since $\tilde{P}^\pi (s,s') = \sum_a \pi(a   \, | \,  s)\mathbb{P}\left(s^{\prime}  \, | \,  s, a\right)$,  $\tilde{P}^\pi $ depends continuously on $\pi$. Matrix inversion is continuous on the set of invertible matrices. 
Hence, both the numerator and denominator in \eqref{eq:Episodicdpi} are continuous in \(\pi\).
Moreover, $\tilde d_0^\top(I-\tilde P^\pi)^{-1}\text{\usefont{U}{bbold}{m}{n}1} \geq\tilde d_0^\top\text{\usefont{U}{bbold}{m}{n}1}=1$. Therefore, \(\tilde{d}_\pi \) is continuous in \(\pi\), and so is \(d_\pi \), since \(d_\pi (z)=0\) for every policy.

It remains to establish the uniform lower bound. The denominator in \eqref{eq:Episodicdpi} is the expected number of transient-state visits before absorption and therefore equals $\mathbb E_{s_0\sim d_0,\pi}[\tau]$. By \cref{lem:absorbing_time}, $\tilde{d}_0 ^\top(I-\tilde{P}^\pi )^{-1}\text{\usefont{U}{bbold}{m}{n}1}=\mathbb E_{s_0\sim d_0,\pi}[\tau]\leq \frac{m}{1-\alpha}$. Moreover,  the expansion $(I-\tilde{P}^\pi )^{-1}=I+\sum_{t=1}^\infty(\tilde{P}^\pi )^t$
implies $\tilde{d}_0 ^\top(I-\tilde{P}^\pi )^{-1}\geq\tilde{d}_0 ^\top$ componentwise.
Therefore, $d_\pi (s)\geq\frac{1-\alpha}{m}d_0(s)$ for every $s\in\tilde{\mathcal S} $.
Consequently, $\inf _\pi \min _{s\in \tilde{\mathcal S} } d_\pi (s) \geq \frac{1-\alpha}{m} \min _{s\in \tilde{\mathcal S} } d_0(s)>0$.
\end{proof}

\begin{lemma}[Uniform positivity of state probabilities in continuing MDPs]
\label{lem:continuing_stationary}
For continuing MDPs, suppose that~\cref{assumption:ergodicity} holds.
Then the stationary distribution $d_\pi$ depends continuously on $\pi$ and satisfies $\inf_{\pi\in\Pi}\min_{s\in\mathcal S}d_\pi(s)>0$.
\end{lemma}
\begin{proof}
For continuing MDPs, $d_\pi$ is the stationary distribution of $P^\pi$.  Under \cref{assumption:ergodicity}, $P^\pi$ is irreducible for every stationary policy $\pi$, and hence admits a unique stationary distribution satisfying $d_\pi(s)>0$ for every state $s\in\mathcal S$. Since $P^\pi(s,s' )= \sum_a \pi(a   \, | \,  s)\mathbb{P}\left(s^{\prime}  \, | \,  s, a\right)$, the transition matrix $P^\pi$ depends continuously on $\pi$. Moreover, for an irreducible finite-state Markov chain, its stationary distribution can be expressed as $d_\pi ^\top = \mathbbm{1}^\top \left(I-P^\pi+\mathbbm{1}\mathbbm{1}^\top\right)^{-1}$. Since the matrix $I-P^\pi+\mathbbm{1}\mathbbm{1}^\top$ is invertible for every $\pi\in\Pi$ and matrix inversion is continuous on the set of invertible matrices, it follows that $d_\pi$ depends continuously on $\pi$.

Because the state and action spaces are finite, the policy space $\Pi:=\prod_{s\in\mathcal S}\Delta(\mathcal A)$ is compact. Therefore, the function $\pi\mapsto \min_{s\in\mathcal S}d_\pi(s)$ is continuous on $\Pi$. It is strictly positive for every $\pi\in\Pi$ and hence attains a strictly positive minimum on the compact set $\Pi$. Consequently, $\inf_{\pi}\min_{s\in\mathcal S}d_\pi (s)>0$. 

\end{proof}

With these uniform positivity results in hand, we now verify the
coverage condition required by~\citep{RL-RT:2021} and conclude the
proof of \cref{thm:convergenceoftabular}.
\begin{proof}
  We apply the convergence framework for tabular off-policy policy
updates developed in~\citep{RL-RT:2021}. For both the direct and softmax parameterizations, \citep[Theorem~2]{RL-RT:2021} establishes global value convergence provided that the state weights $d_t(s)$ satisfy the cumulative coverage condition
\begin{equation}\label{eq:condition_convergence}
    \sum_{t=0}^{\infty}\eta_t d_t(s)=\infty,
    \quad \forall s\in\mathcal S.
\end{equation}
We show that the state weights induced by our biased updates satisfy
this condition.

Let $\pi_t:=\pi_{\theta_t}$. Our biased
updates correspond to the framework of~\citep{RL-RT:2021} with $d_t=d_{\pi_t}$ and $\eta_t=\eta$, 
where $\eta>0$ is constant. For continuing MDPs, \cref{lem:continuing_stationary}
ensures that $\inf_{\pi\in\Pi}d_\pi(s)>0$ for all $s$. Thus, for any constant stepsize $\eta>0$, the state distribution satisfies $ \sum_{t=0}^{\infty}\eta d_{\pi_t}(s)=\infty$ and thus \eqref{eq:condition_convergence} holds.

For episodic MDPs, \cref{cor:episodic_uniform_bound} similarly gives $\inf_{\pi\in\Pi}d_\pi(s)>0$ 
for every $s\in\tilde{\mathcal S} $. The only state not covered by this bound is the absorbing state $z$,
for which $d_\pi(z)=0$. To place the episodic update within the
framework of~\citep{RL-RT:2021}, define
the auxiliary state weights
\[
     w_\pi(s)=
    \begin{cases}
        d_\pi(s), & s\ne z,\\
        1, & s=z.
    \end{cases}
\]
The framework in~\citep[Appendix~A.4]{RL-RT:2021} permits arbitrary
nonnegative state weights and does not require them to be normalized. 
Replacing $d_\pi$ by $w_\pi$
does not alter the update direction. Indeed, the two weight functions differ only at the absorbing state $z$, where
$Q^\pi(z,a)=0$ for every action $a\in\mathcal A$. Consequently,  $\sum_{s,a}w_\pi(s)Q^\pi(s,a)
\nabla_\theta\pi(a\,|\, s)=\sum_{s,a}d_\pi(s)Q^\pi(s,a)\nabla_\theta\pi(a\,|\, s).$ Thus, $w_\pi$ and $d_\pi$ induce identical updates 
under both tabular parameterizations.  In the direct case, the same projection therefore
also produces identical iterates. Moreover, the uniform lower bound on $\tilde{\mathcal S} $ and
$w_{\pi_t}(z)=1$ imply that $ \sum_{t=0}^{\infty}\eta w_{\pi_t}(s)=\infty$ for every $s\in\mathcal S$. Therefore, the equivalent update also satisfies  \eqref{eq:condition_convergence} in  the episodic setting as well.

Thus, in both episodic and continuing MDPs, the conditions required by 
\citep[Theorem~2 and its proof]{RL-RT:2021} are satisfied. It follows that 
$V^{\pi_t}(s)\to V^*(s)$ for every state $s\in\mathcal S$.
Since $\mathcal S$ is finite, $\lim_{t\to\infty}J(\pi_t)
    =\lim_{t\to\infty}\sum_{s\in\mathcal S}
      d_0(s)V^{\pi_t}(s)
    =J^*$. 
This completes the proof.
\end{proof}
}

\subsection{\texorpdfstring
{Proof of \cref{thm:softmax_asymptotic}}
{Proof of the exact asymptotic rates}}
\label{prf:softmax_asymptotic}

\begin{proof}
If $|\mathcal A|=1$, the unique policy is optimal and both conclusions hold trivially.
Henceforth, suppose $|\mathcal A|\geq2$.

We analyze the normalized exact and biased updates in parallel. For either update, write
\begin{equation}\label{eq:asymptotic_update}
    \theta_{t+1,s,a}
    =
    \theta_{t,s,a}
    +\eta w_t(s)\pi_t(a\, | \,  s)A^{\pi_t}(s,a),
\end{equation}
where $w_t(s)=d_{\pi_t,\gamma}(s)$ for the normalized exact update and $w_t(s)=d_{\pi_t}(s)$ for the biased update.

We first establish the convergence of the advantage functions $A^{\pi_t}$, policies $\pi_t$,  and state weights $w_{t}$ needed for the asymptotic analysis.
In episodic MDPs, the absorbing state $z$ can be omitted,
since $A^{\pi_t}(z,a)=0$ and $w_t(z)=0$ for every $t$ and $a$.
We therefore work on the relevant state set $\tilde{\mathcal S} $. For the biased update, \cref{thm:convergenceoftabular} gives $V^{\pi_t}(s) \rightarrow V^*(s)$ and $J\left(\pi_t\right) \rightarrow J^*$. For the normalized exact update, the discounted occupancy measure satisfies $\mu_{\pi, \gamma}(s) \geq d_0(s)$, while $
\kappa_\pi \leq\sum_{s\in\mathcal S}\mu_{\pi,\gamma}(s) =\frac{1}{1-\gamma}$.  Hence, for every $s\in\tilde{\mathcal S} $, $ d_{\pi,\gamma} (s)
    =\frac{\mu_{\pi,\gamma}(s)}{\kappa_\pi} \geq (1-\gamma)d_0(s)>0$. 
   Thus, the normalized exact update satisfies the same coverage condition used in the proof of \cref{thm:convergenceoftabular} in \cref{prf:convergenceoftabular}, and the same argument  yields $V^{\pi_t}(s) \rightarrow V^*(s)$ and \(J(\pi_t)\rightarrow J^*\) for this update as well.
   Since $\mathcal{S}$ is finite, the Bellman equation then gives, for both updates, $Q^{\pi_t}(s, a)=R(s, a)+\gamma \sum \mathbb{P}\left(s^{\prime} \, | \,  s, a\right) V^{\pi_t}\left(s^{\prime}\right) \rightarrow Q^*(s,a)$ and consequently, $A^{\pi_t}(s, a) \rightarrow A^*(s,a):=Q^*(s, a)-V^*(s)=A^{\pi^*}(s, a)$.
    For each \(s\in\tilde{\mathcal S} \) and \(a\neq a^*(s)\), define 
    \[ \delta_{s,a}:=V^*(s)-Q^*(s,a)>0, \]
    where strict positivity follows from   \cref{assumption:unique_optimal_action}. 
    Hence,
    $$
    A^{*}(s, a)=Q^*(s, a)-V^*(s)=\left\{\begin{array}{ll}
0, & a=a^*(s), \\
-\delta_{s, a}, & a \neq a^*(s).
\end{array} \right.
    $$
    By \cref{lemma:performance_difference}
    \begin{equation}\label{eq:objective_gap}
        \begin{aligned}
    J^*-J(\pi_t) &= -\sum_{s \in \tilde{\mathcal S} } \mu_{\pi_t, \gamma}(s) \sum_{a \in \mathcal{A}} \pi_t(a \, | \,  s) A^{*}(s, a)  \\ & = \sum_{s\in\tilde{\mathcal S} } \mu_{\pi_t,\gamma}(s) \sum_{a\neq a^*(s)} \pi_t(a\, | \,  s)\delta_{s,a}.       
    \end{aligned}
    \end{equation}
  Since \(\mu_{\pi_t,\gamma}(s)\geq d_0(s)>0\), \(\pi_t(a\, | \,  s)\geq0\), and \(\delta_{s,a}>0\) for every \(s\in\tilde{\mathcal S} \) and \(a\neq a^*(s)\),  we have
    \(J^*-J(\pi_t) \geq
d_0(s)\pi_t(a\, | \,  s)\delta_{s,a} \geq 0
.\) Therefore, 
$$0\leq\pi_t(a\, | \,  s) \leq \frac{J^*-J(\pi_t)}{d_0(s)\delta_{s,a}} \longrightarrow0.$$
Since $\mathcal{A}$ is finite, 
$$
\pi_t\left(a^*(s) \, | \,  s\right)=1-\sum_{a \neq a^*(s)} \pi_t(a \, | \,  s) \longrightarrow 1 .
$$
Thus, $\pi_t \longrightarrow \pi^* $ on $\tilde{\mathcal S} $. We next establish the convergence of the discounted state distributions. The continuity and positivity of \(d_\pi\) established in \cref{cor:episodic_uniform_bound,lem:continuing_stationary} imply that for all $s \in \tilde{\mathcal S} $,  \(d_{\pi_t}(s)\longrightarrow d_{\pi^*}(s)>0\). Moreover, since $\mu_{\pi, \gamma}^{\top}=\sum_{k=0}^{\infty} \gamma^k d_0^{\top}\left(P^\pi\right)^k=d_0^{\top}\left(I-\gamma P^\pi\right)^{-1}$, \(P^\pi\) depends continuously on \(\pi\), and \(I-\gamma P^\pi\) is invertible for \(\gamma<1\), \(\mu_{\pi,\gamma}\) depends continuously on \(\pi\).  Since
$\kappa_\pi=\sum_{s\in\tilde{\mathcal S} }\mu_{\pi,\gamma}(s)$ is a finite sum of continuous functions, the normalized distribution \(d_{\pi,\gamma}(s)=\mu_{\pi,\gamma}(s)/\kappa_\pi\) is also continuous in \(\pi\). Therefore, $d_{\pi_t, \gamma}(s) \longrightarrow d_{\pi^*, \gamma}(s)>0$. %
Consequently, for $s \in \tilde{\mathcal S} $, $w_t(s) \longrightarrow w_*(s)>0$, where $w_*(s) = d_{\pi^*, \gamma}(s)$ for the normalized exact update and $w_*(s) = d_{\pi^*}(s)$ for the biased update.

We next determine the asymptotic probability ratios $\lim_{t\to \infty}\frac{\pi_t(b \, | \, s)}{\pi_t(a \, | \, s)}$ between suboptimal actions $a,b\neq a^*(s)$. Fix $s\in \tilde{\mathcal S} $ %
and define $Z_t:=\sum_{b \in \mathcal{A}} e^{\theta_{t, s, b}}$. 
The softmax parameterization gives $\pi_t(a\, | \,  s)e^{-\theta_{t,s,a}}=\frac{1}{Z_t}$.
Since for every \(a\neq a^*(s)\), $\pi_t(a\, | \,  s)\to0$, $A^{\pi_t}(s,a)\to-\delta_{s,a}$, and $w_t(s)\to w_*(s)$, we have
$$
\theta_{t+1, s, a}-\theta_{t, s, a}=\eta w_t(s) \pi_t(a \, | \,  s) A^{\pi_t}(s, a) \longrightarrow 0 .
$$
Moreover, since \(A^{\pi_t}(s,a)\to-\delta_{s,a}<0\) for every $a\neq a^*(s)$ and $\mathcal{A}$ is finite, there exists $T_1$ such that for all $t\geq T_1$ and $a\neq a^*(s)$,  we have $A^{\pi_t}(s,a)<0$.
Therefore, $\theta_{t+1, s, a}-\theta_{t, s, a}<0$ for all  $a\neq a^*(s)$ and $t\geq T_1$, and consequently, \(e^{-\theta_{t,s,a}}\) is  strictly increasing for $t\geq T_1$. 
Using $e^{-u} - 1 = -u(1+o(1))$ as \(u\to0\), we obtain
\begin{equation}\label{eq:exp-exp}
    \begin{aligned}
& \quad e^{-\theta_{t+1, s, a}}-e^{-\theta_{t, s, a}} \\& =e^{-\theta_{t, s, a}}\left[e^{-\eta w_t(s) \pi_t(a \, | \, s) A^{\pi_t}(s, a)}-1\right] \\
& =-\eta w_t(s) \pi_t(a \, | \, s)A^{\pi_t}(s, a)  e^{-\theta_{t, s, a}}(1+o(1)) \\
& =-\frac{\eta w_t(s) A^{\pi_t}(s, a)}{Z_t}(1+o(1)) \\
& =\frac{\eta w_t(s)}{Z_t}\left(\delta_{s, a}+o(1)\right).
\end{aligned}
\end{equation}
Here and below, \(o(1)\) denotes a quantity tending to zero as \(t\to\infty\).  
For any two suboptimal actions $a,b\neq a^*(s)$, \eqref{eq:exp-exp} gives
\begin{equation}\label{eq:ratio}
    \frac{e^{-\theta_{t+1,s,a}}-e^{-\theta_{t,s,a}}
}{e^{-\theta_{t+1,s,b}}-e^{-\theta_{t,s,b}}}=
\frac{\delta_{s,a}+o(1)}{\delta_{s,b}+o(1)}
\longrightarrow
\frac{\delta_{s,a}}{\delta_{s,b}}.
\end{equation}
We next show that $e^{-\theta_{t, s, a}} \rightarrow \infty$ for every \(a\neq a^*(s)\). Suppose, to the contrary, that $e^{-{\theta_{t,s,b}}}$ is bounded for some suboptimal action $b$. By \eqref{eq:ratio}, there exists $T\geq T_1$ such that for every $a\neq a^*(s)$ and $t\geq T$,
$$0<e^{-\theta_{t+1, s, a}}-e^{-\theta_{t, s, a}} \leq \frac{2 \delta_{s, a}}{\delta_{s, b}}\left(e^{-\theta_{t+1, s, b}}-e^{-\theta_{t, s, b}}\right) .$$
Summing over $t$ from $T$ to $t'-1$ gives
\begin{equation*}\begin{adjustbox}{max width=\columnwidth}$\displaystyle
e^{-\theta_{T, s, a}}< e^{-\theta_{t', s, a}} \leq  e^{-\theta_{T, s, a}}  +\frac{2 \delta_{s, a}}{\delta_{s, b}}\left(e^{-\theta_{t', s, b}}-e^{-\theta_{T, s, b}}\right).
$\end{adjustbox}\end{equation*}
The right-hand side is bounded because $e^{-\theta_{t', s, b}}$ is bounded by assumption and \(s,a,b,T\) are fixed. Thus, $e^{-\theta_{t', s, a}}$ is bounded above and bounded away from zero for every $a\neq a^*(s)$. Consequently, every suboptimal logit \(\theta_{t,s,a}\) is bounded  above and below.
On the other hand, since $\sum_a \pi_t(a \, | \, s) A^{\pi_t}(s, a)=0$, \eqref{eq:asymptotic_update} implies
\begin{equation}\begin{adjustbox}{max width=\dimexpr\columnwidth-3em\relax}$\displaystyle \label{eq:constant}
    \sum_a\left(\theta_{t+1, s, a}-\theta_{t, s, a}\right)=\eta w_t(s) \sum_a \pi_t(a \, | \, s) A^{\pi_t}(s, a)=0.
$\end{adjustbox}\end{equation}
Thus $\sum_a \theta_{t, s, a}=\sum_a \theta_{0, s, a}$ for all $t$. Since all suboptimal logits are bounded and the sum of all logits is constant, the optimal logit is also bounded.
Hence all logits at state \(s\) are bounded, which implies that all softmax  probabilities of the suboptimal actions are bounded away from zero. This  contradicts \(\pi_t(a^*(s)\, | \, s)\to1\).
Therefore, \(e^{-\theta_{t,s,a}}\) is unbounded for every suboptimal action \(a\neq a^*(s)\). Since \(e^{-\theta_{t,s,a}}\) is eventually strictly increasing, it follows that $e^{-\theta_{t,s,a}}\longrightarrow\infty$ for all  $a\neq a^*(s)$. We now apply the Stolz–Ces{à}ro theorem to \eqref{eq:ratio}, which gives
$\frac{e^{-\theta_{t,s,a}}}{e^{-\theta_{t,s,b}}}
\longrightarrow
\frac{\delta_{s,a}}{\delta_{s,b}}$ for all $a,b\neq a^*(s)$. By the softmax parametrization, we have
\begin{equation} \label{eq:ratioofpi}
    \frac{\pi_t(b \, | \, s)}{\pi_t(a \, | \, s)}=\frac{e^{\theta_{t, s, b}}}{e^{\theta_{t, s, a}}}=\frac{e^{-\theta_{t, s, a}}}{e^{-\theta_{t, s, b}}}=\frac{\delta_{s, a}}{\delta_{s, b}}+o(1).
\end{equation}

We next determine the limit of \(t\pi_t(a\, | \, s)\) for each suboptimal action \(a\neq a^*(s)\).
Fix $a\neq a^*(s)$. From~\eqref{eq:constant}, 
\begin{equation*}\begin{adjustbox}{max width=\columnwidth}$\displaystyle
\begin{aligned}
\theta_{t+1, s, a^*(s)}-\theta_{t, s, a^*(s)} & =-\sum_{b \neq a^*(s)}\left(\theta_{t+1, s, b}-\theta_{t, s, b}\right) \\
& =-\eta w_t(s) \sum_{b \neq a^*(s)} \pi_t(b \, | \, s) A^{\pi_t}(s, b) .
\end{aligned}
$\end{adjustbox}\end{equation*}
Subtracting the update of \(\theta_{t,s,a}\) gives
\begin{equation}\begin{adjustbox}{max width=\dimexpr\columnwidth-3em\relax}$\displaystyle  \label{eq:theta-theta}
    \begin{aligned}
&\quad \left(\theta_{t+1, s, a^*(s)}-\theta_{t+1, s, a}\right)-\left(\theta_{t, s, a^*(s)}-\theta_{t, s, a}\right) \\
& =-\eta w_t(s) \Bigl[  \sum_{b \neq a^*(s)} \pi_t(b \, | \, s) A^{\pi_t}(s, b)+\pi_t(a \, | \, s) A^{\pi_t}(s, a)\Bigr] \\
& =-\eta w_t(s) \pi_t(a \, | \, s)\Bigl[\sum_{b \neq a^*(s)} \frac{\pi_t(b \, | \, s)}{\pi_t(a \, | \, s)} A^{\pi_t}(s, b)+A^{\pi_t}(s, a)\Bigr].
\end{aligned}
$\end{adjustbox}\end{equation}
For every \(b\neq a^*(s)\), \eqref{eq:ratioofpi} and $-A^{\pi_t}(s, b)=\delta_{s, b}+o(1)$ imply 
\begin{equation}\label{eq:ratioandA}
    \begin{aligned}
-\frac{\pi_t(b \, | \, s)}{\pi_t(a \, | \, s)} A^{\pi_t}(s, b) & =-\left(\frac{\delta_{s, a}}{\delta_{s, b}}+o(1)\right)\left(-\delta_{s, b}+o(1)\right) \\
& =\delta_{s, a}+o(1). 
\end{aligned}
\end{equation}
Substituting  \eqref{eq:ratioandA} and $-A^{\pi_t}(s, a)=\delta_{s, a}+o(1)$ into \eqref{eq:theta-theta} yields
\begin{equation} \label{eq:differnecetheta}
    \begin{aligned}
&\quad \left(\theta_{t+1, s, a^*(s)}-\theta_{t+1, s, a}\right)-\left(\theta_{t, s, a^*(s)}-\theta_{t, s, a}\right) \\
& =\eta w_t(s) \pi_t(a \, | \, s)\left[(|\mathcal{A}|-1) \delta_{s, a}+\delta_{s, a}+o(1)\right] \\
& =\eta w_t(s) \pi_t(a \, | \, s)\left(|\mathcal{A}| \delta_{s, a}+o(1)\right) .
\end{aligned}
\end{equation}
We next study the ratio $\frac{\pi_t\left(a^*(s) \, | \, s\right)}{\pi_t(a \, | \, s)}=e^{\theta_{t, s, a^*(s)}-\theta_{t, s, a}}$. Using~\eqref{eq:differnecetheta} and $e^{u} - 1 = u(1+o(1))$ as \(u\to0\), we obtain
\begin{equation}\label{eq:explimit}
    \begin{aligned}
&\quad e^{\theta_{t+1, s, a^*(s)}-\theta_{t+1, s, a}}-e^{\theta_{t, s, a^*(s)}-\theta_{t, s, a}} \\
& =e^{\theta_{t, s, a^*(s)}-\theta_{t, s, a}}\left[e^{\eta w_t(s) \pi_t(a \, | \, s)\left(|\mathcal{A}| \delta_{s, a}+o(1)\right)}-1\right] \\
& =e^{\theta_{t, s, a^*(s)}-\theta_{t, s, a}} \eta w_t(s) \pi_t(a \, | \, s)\left(|\mathcal{A}| \delta_{s, a}+o(1)\right) \\
& =\eta w_t(s) \pi_t\left(a^*(s) \, | \, s\right)\left(|\mathcal{A}| \delta_{s, a}+o(1)\right)\\
& \rightarrow  \eta w_*(s)|\mathcal{A}| \delta_{s, a}>0.
\end{aligned}
\end{equation}
Since $e^{\theta_{t+1, s, a^*(s)}-\theta_{t+1, s, a}}-e^{\theta_{t, s, a^*(s)}-\theta_{t, s, a}}$ can be rewritten as $\frac{e^{\theta_{t+1, s, a^*(s)}-\theta_{t+1, s, a}}-e^{\theta_{t, s, a^*(s)}-\theta_{t, s, a}}}{(t+1)-t}$, whose denominator sequence $t$ is strictly increasing and unbounded, the Stolz–Ces{à}ro theorem gives
\begin{equation*}%
    \frac{e^{\theta_{t, s, a^*(s)}-\theta_{t, s, a}}}{t} \longrightarrow \eta w_*(s)|\mathcal{A}| \delta_{s, a}.
\end{equation*}
By the softmax parameterization, $t \pi_t(a \, | \, s)=\frac{\pi_t\left(a^*(s) \, | \, s\right)}{e^{\theta_{t, s, a^*(s)}-\theta_{t, s, a}} / t}$, and hence
\begin{equation}\label{eq:action_probability_rate}
    t \pi_t(a \, | \, s) \longrightarrow \frac{1}{\eta w_*(s)|\mathcal{A}| \delta_{s, a}}, \textup{ for }a \neq a^*(s) .
\end{equation}

Finally, using~\eqref{eq:objective_gap},  \(\mu_{\pi_t,\gamma}(s)\to\mu_{\pi^*,\gamma}(s)\), and the finiteness of
$\tilde{\mathcal S} $ and $\mathcal A$, we take limits termwise to obtain
\begin{equation*}\begin{adjustbox}{max width=\columnwidth}$\displaystyle
\begin{aligned}
\lim _{t \rightarrow \infty} t\left[J^*-J\left(\pi_t\right)\right] & =\sum_{s \in \tilde{\mathcal S} } \mu_{\pi^*, \gamma}(s) \sum_{a \neq a^*(s)} \delta_{s, a} \lim _{t \rightarrow \infty} t \pi_t(a \, | \, s) \\
& =\sum_{s \in \tilde{\mathcal S} } \mu_{\pi^*, \gamma}(s) \sum_{a \neq a^*(s)} \frac{1}{\eta w_*(s)|\mathcal{A}|} \\
& =\frac{|\mathcal{A}|-1}{\eta|\mathcal{A}|} \sum_{s \in \tilde{\mathcal S} } \frac{\mu_{\pi^*, \gamma}(s)}{w_*(s)}.
\end{aligned}
$\end{adjustbox}\end{equation*}
For the normalized exact update, \(w_*(s)=d_{\pi^*,\gamma}(s)\) and \(\mu_{\pi^*,\gamma}(s)=\kappa_{\pi^*}d_{\pi^*,\gamma}(s)\), which gives
$$
\lim _{t \rightarrow \infty} t\left[J^*-J\left(\pi_t^{\mathrm{e}}\right)\right]=\frac{\kappa_{\pi^*}(|\mathcal{A}|-1)|\tilde{\mathcal S} |}{\eta|\mathcal{A}|} .
$$
For the biased update, \(w_*(s)=d_{\pi^*}(s)\), and hence
\[
\lim_{t\to\infty}t[J^*-J(\pi_t^{\mathrm b})]
=
\frac{\kappa_{\pi^*}(|\mathcal A|-1)}
{\eta|\mathcal A|}
\sum_{s\in\tilde{\mathcal S} }
\frac{d_{\pi^*,\gamma}(s)}{d_{\pi^*}(s)}.
\]
This completes the proof.

\end{proof}

\subsection{An example where the biased gradient achieves faster convergence under general parametrization}\label{example:biased}
We consider the gridworld example from \citep[Example 4.1]{RS-AB:2018}, which is also employed in \cref{sec:simulation}. The gridworld is a $4 \times 4$ grid, where each cell represents a state, with $\mathcal{S} = \{1, \dots, 16\}$, and states $1$ and $16$ are terminal states. In each state, the agent can take one of four possible actions: $\mathcal{A} = \{\text{up}, \text{down}, \text{left}, \text{right}\}$. In the environment,  a northwest wind is present, such that regardless of the agent’s current state and chosen action, there is an additional $\epsilon$ probability of transitioning to the right and  $\epsilon$ probability to the  downward state, and a probability of $1 - 2\epsilon$ to move according to the intended action. The reward for all transitions is $-1$ until the terminal state is reached. The objective is to find the shortest path to one of the terminal states. We take $\epsilon=0.1$ and for all nonterminal states $s$, the policy is $\pi(\textup{up} \,|\, s) =\pi(\textup{right} \,|\, s)=\frac{e^\theta}{2(1+e^\theta)}$ and $\pi(\textup{down} \,|\, s) =\pi(\textup{left} \,|\, s)=\frac{1}{2(1+e^\theta)}$. 
\begin{figure}[htbp] 
\centering    \includegraphics[width=0.24\textwidth]{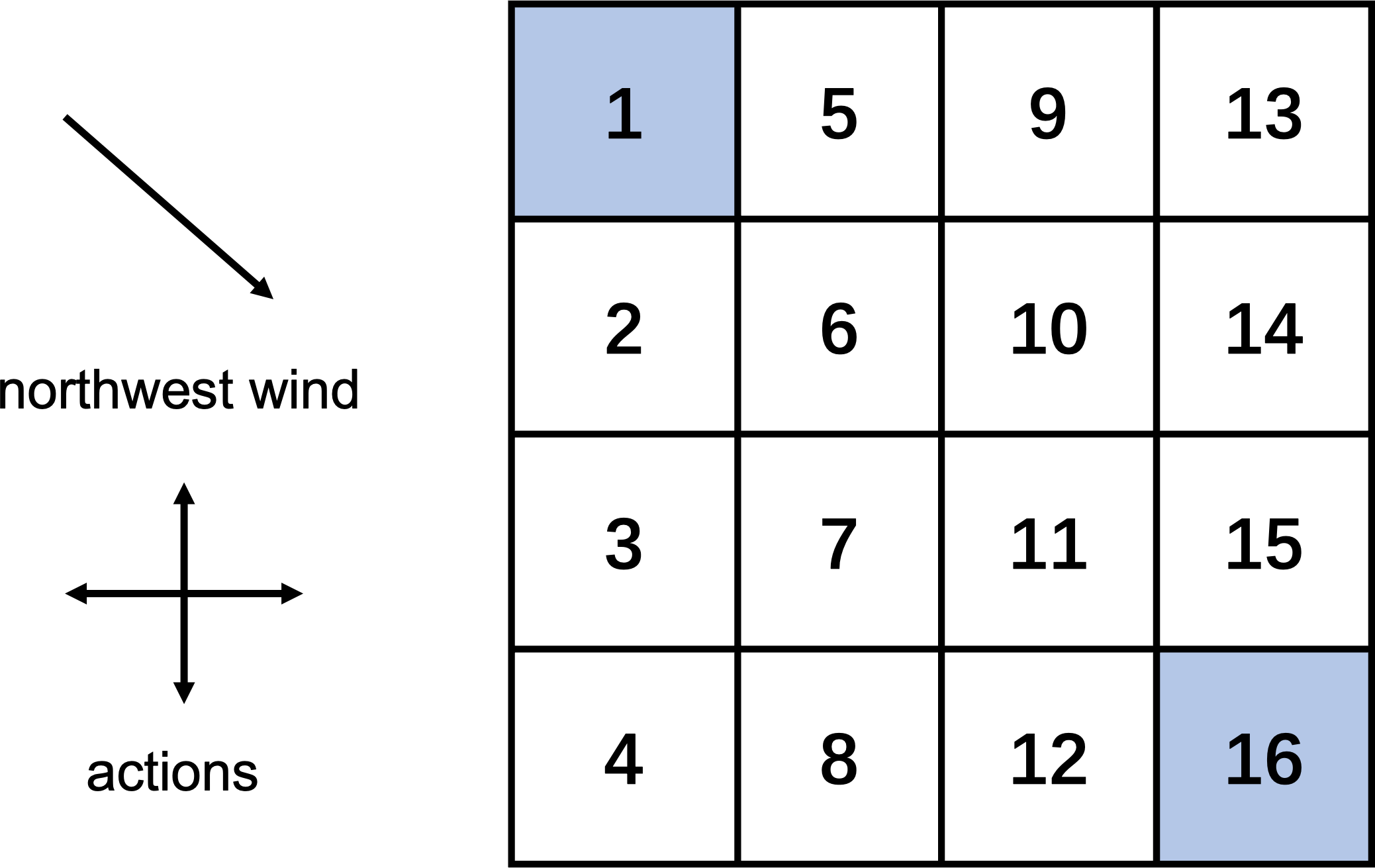}
\caption{The $4\times4$ gridworld with terminal states 1 and 16 and stochastic wind toward the right and down.}
\label{fig:gridworld}
\end{figure}

\begin{figure*}[!htbp] 
\centering
\includegraphics[width=\textwidth]{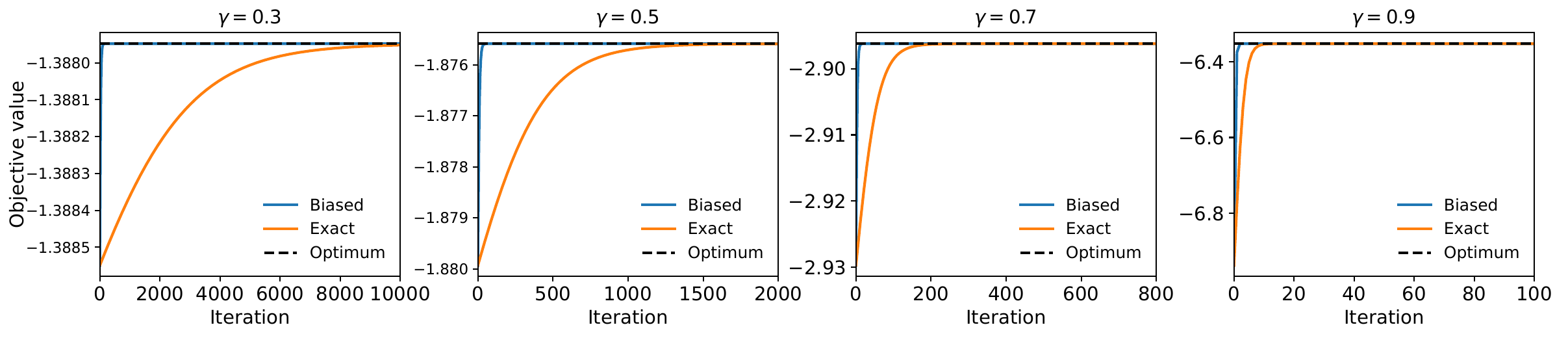}
\caption{Objective values under biased and unbiased policy-gradient updates with a shared scalar policy parameter in the gridworld task.}
\label{figure:general}
\end{figure*}

We employ the unbiased policy gradient  update $ \theta_{t+1} =\theta_{t}+\eta \nabla_\theta^{\textup{e}} J (\theta_t)$ and the biased policy gradient  update $\theta_{t+1} =\theta_{t}+\eta \hat{\nabla}^{\textup{e}}_\theta J (\theta_t)$  for varying discount factor $\gamma$. The results are shown in \cref{figure:general}, \textcolor{black}{where the reference optimum within the restricted policy class is computed numerically by evaluating $J(\theta)$ through the Bellman equations and maximizing it over the shared scalar parameter \(\theta\)}. 
From  the experimental results, we observe that under this restricted parameterization (the policy class is restricted such that not all optimal or possible policies can be represented), the biased algorithm not only converges to the same \textcolor{black}{optimal value} as the unbiased method but also exhibits faster convergence. As  $\gamma$  approaches $1$, the distribution mismatch diminishes, and the performance of the two algorithms becomes more similar.

This simple example demonstrates that despite the existence of distribution shift and the resulting theoretical gap, the biased policy gradient method can still perform well, and even outperform the unbiased method in practice. 
Experiments reported in \citep{SZ-RL-VH:2022, FC-MG-AM:2023, HP-DY-MS:2023} also indicate that correcting the distribution mismatch does not always improve empirical performance.

\section{\texorpdfstring{Proof of \cref{subsec:bounds}}{Proof: subsec bounds}} \label{appendix:bounds}
\subsection{\texorpdfstring{Proof of \cref{thm:boundforepisodic}}{Proof: thm boundforepisodic}} \label{prf:boundforepisodic}
\begin{proof}
Recall the transient-state transition matrix $\tilde{P}^\pi $ and the restricted distributions $\tilde{d}_0 $ and $\tilde{d}_\pi $ introduced in the proof of \cref{cor:episodic_uniform_bound}.
Similarly, let $\tilde{d}_{\pi,\gamma} $ denote the restriction of $d_{\pi,\gamma} $ to $\tilde{\mathcal S} $. Since $d_{\pi,\gamma} (z)=d_\pi (z)=0$, $\|d_{\pi,\gamma} -d_\pi \|_1=\|\tilde{d}_{\pi,\gamma} -\tilde{d}_\pi \|_1$.
It therefore suffices to work with these
restricted distributions. %
By \cref{lemma:absorbing}, $(\tilde{P}^\pi )^m\text{\usefont{U}{bbold}{m}{n}1}\leq\alpha\text{\usefont{U}{bbold}{m}{n}1}$.
Since $\tilde{P}^\pi $ is row-substochastic, for every $t\geq0$, $(\tilde{P}^\pi )^t\text{\usefont{U}{bbold}{m}{n}1}\leq\text{\usefont{U}{bbold}{m}{n}1}$. Hence, 
\begin{equation}\label{eq:bounddpi}
	\begin{aligned}
		(I-\tilde{P}^\pi )^{-1} \text{\usefont{U}{bbold}{m}{n}1} = \sum_{t=0}^\infty (\tilde{P}^\pi )^t \text{\usefont{U}{bbold}{m}{n}1} &= \sum_{k=0}^\infty \sum_{t=0}^{m-1}  (\tilde{P}^\pi )^{km+t}  \text{\usefont{U}{bbold}{m}{n}1} \\
		& \leq  \sum_{k=0}^\infty \sum_{t=0}^{m-1}  (\tilde{P}^\pi )^{km} \text{\usefont{U}{bbold}{m}{n}1} \\
		&=  m\sum_{k=0}^\infty  (\tilde{P}^\pi )^{km}  \text{\usefont{U}{bbold}{m}{n}1}\\
		&\leq m \sum_{k=0}^\infty \alpha^k \text{\usefont{U}{bbold}{m}{n}1}  = \frac{m}{1-\alpha} \text{\usefont{U}{bbold}{m}{n}1}.
	\end{aligned}
\end{equation}
Similar derivation gives 
\begin{equation}\label{eq:bounddpigamma}
	(I- \gamma\tilde{P}^\pi )^{-1} \text{\usefont{U}{bbold}{m}{n}1} \leq \frac{m}{1-\alpha \gamma^m} \text{\usefont{U}{bbold}{m}{n}1}.
\end{equation}

Let $Y:=(I-\tilde{P}^\pi )^{-1}$ and  $Y_\gamma:=(I-\gamma\tilde{P}^\pi )^{-1}$. The normalized undiscounted and discounted state distributions are $\tilde{d}_\pi ^\top = \frac{\tilde{d}_0 ^\top Y} {\tilde{d}_0 ^\top Y\text{\usefont{U}{bbold}{m}{n}1}}$ and $\tilde{d}_{\pi,\gamma} ^\top = \frac{\tilde{d}_0 ^\top Y_\gamma} {\tilde{d}_0 ^\top Y_\gamma\text{\usefont{U}{bbold}{m}{n}1}}$, respectively. Using the  Woodbury matrix identity, we have 
\begin{equation}\label{eq:Woodbury_new}
	Y_\gamma
	=
	Y-(1-\gamma)Y\tilde{P}^\pi Y_\gamma.    
\end{equation}
Define $q_\gamma^\top := \tilde{d}_0 ^\top Y\tilde{P}^\pi Y_\gamma$, and let $Z:=\tilde{d}_0 ^\top Y\text{\usefont{U}{bbold}{m}{n}1}$ and $ Z_\gamma:=\tilde{d}_0 ^\top Y_\gamma\text{\usefont{U}{bbold}{m}{n}1}$. By \eqref{eq:kappa}, $Z_\gamma=\kappa_\pi$. Multiplying from the left $\tilde{d}_0 ^\top$ on both sides of~\eqref{eq:Woodbury_new} yields 
\begin{equation}\label{eq:d0Y}
	\tilde{d}_0 ^\top Y_\gamma
	=
	\tilde{d}_0 ^\top Y-(1-\gamma)q_\gamma^\top.
\end{equation}
Further multiplying from the right $\text{\usefont{U}{bbold}{m}{n}1}$ on both sides of~\eqref{eq:d0Y} yields 
\begin{equation}\label{eq:d0Y1}
	Z_\gamma
	=
	Z-(1-\gamma)q_\gamma^\top\text{\usefont{U}{bbold}{m}{n}1}.
\end{equation}
Then,
\begin{align}\label{eq:differenceofdistribution}
	\begin{split}
		\tilde{d}_{\pi,\gamma} ^\top-\tilde{d}_\pi ^\top
		&=
		\frac{\tilde{d}_0 ^\top Y-(1-\gamma)q_\gamma^\top}{Z_\gamma}
		-
		\frac{\tilde{d}_0 ^\top Y}{Z}\\
		&=
		(1-\gamma)
		\frac{
			(q_\gamma^\top\text{\usefont{U}{bbold}{m}{n}1})\tilde{d}_0 ^\top Y
			-
			Zq_\gamma^\top
		}{
			ZZ_\gamma
		},    
	\end{split}
\end{align}
where we used~\eqref{eq:d0Y} in the first equality and~\eqref{eq:d0Y1} in the second equality. Since all the vectors involved are nonnegative, we have $\|\tilde{d}_0 ^\top Y\|_1=Z$ and $\|q_\gamma^\top\|_1=q_\gamma^\top\text{\usefont{U}{bbold}{m}{n}1}$. Moreover, $Z_\gamma\geq1$, because $Y_\gamma=I+\gamma\tilde{P}^\pi +\gamma^2(\tilde{P}^\pi )^2+\cdots$ and $\tilde{d}_0 ^\top\text{\usefont{U}{bbold}{m}{n}1}=1$. Hence, taking $1$-norm on both sides of~\eqref{eq:differenceofdistribution} yields
\[
\begin{aligned}
	\|\tilde{d}_{\pi,\gamma} -\tilde{d}_\pi \|_1
	&\leq
	2(1-\gamma)
	\frac{q_\gamma^\top\text{\usefont{U}{bbold}{m}{n}1}}{Z_\gamma}.
\end{aligned}
\]
Since $Y$ and $Y_\gamma$ are convergent power series in $\tilde{P}^\pi $, these three matrices commute, i.e., $Y \tilde{P}^\pi Y_\gamma=Y_\gamma \tilde{P}^\pi Y$. Hence, $q_\gamma^\top=\tilde{d}_0 ^\top Y_\gamma\tilde{P}^\pi Y$.
By~\eqref{eq:bounddpi} and the row-substochasticity of $\tilde{P}^\pi $,
\[
\tilde{P}^\pi Y\text{\usefont{U}{bbold}{m}{n}1}
\leq \frac{m}{1-\alpha}\tilde{P}^\pi \text{\usefont{U}{bbold}{m}{n}1}
\leq \frac{m}{1-\alpha}\text{\usefont{U}{bbold}{m}{n}1}.
\]
Multiplying by the nonnegative row vector $\tilde{d}_0 ^\top Y_\gamma$ yields
\[
q_\gamma^\top\text{\usefont{U}{bbold}{m}{n}1}
=\tilde{d}_0 ^\top Y_\gamma\tilde{P}^\pi Y\text{\usefont{U}{bbold}{m}{n}1}
\leq \frac{m}{1-\alpha}Z_\gamma.
\]
Consequently,
\[
\|d_{\pi,\gamma} -d_\pi \|_1
=\|\tilde{d}_{\pi,\gamma} -\tilde{d}_\pi \|_1
\leq \frac{2m}{1-\alpha}(1-\gamma).
\]
This completes the proof of \cref{thm:boundforepisodic}.
\end{proof}

\section{\texorpdfstring{Proof of \cref{subsec:biasedgradient}}{Proof: subsec biasedgradient}}  \label{appendix:biasedgradient}
\subsection{\texorpdfstring{Proof of \cref{thm:gradbias_ep}}{Proof: thm gradbias ep}}  \label{proof:gradbias_ep}
\begin{proof}
We first establish a uniform bound on the state-action value function $Q^\pi$. Fix $s\in\tilde{\mathcal S} $ and $a\in\mathcal A$. Let $\pi'$ be the stationary policy that selects $a$ with probability one at $s$ and agrees with $\pi$ at every other state. Define $\sigma:=\inf \left\{t \geq 1: s_t \in\{s, z\}\right\}$, i.e., the first time after leaving the initial state that the process either returns to $s$ or reaches the absorbing state $z$. Let $p:=\mathbb P_{\pi'}(s_\sigma=s\,|\, s_0=s)$, so that $p$ is the probability of returning to $s$ before absorption, while $1-p$ is the probability of reaching $z$ before returning to $s$. Consider two processes starting from $s$. The first selects action $a$ at time $0$ and follows $\pi$ thereafter, while the second follows $\pi'$ throughout. By construction, $\pi'$ selects $a$ at $s$ and agrees with $\pi$ at every state other than $s$. Hence, before time $\sigma$, the two processes evolve according to the same transition law. In particular, they have the same distribution of $\sigma$ and the same probability $p$ of returning to $s$ before reaching $z$. We now decompose the absorption time at $\sigma$. For the first process, if $s_\sigma=z$, the process is already absorbed and the remaining time is zero. If $s_\sigma=s$, which occurs with probability $p$, then from time $\sigma$ onward the process follows $\pi$ starting again from state $s$. By the strong Markov property, the conditional expected remaining absorption time is therefore $\mathbb E_\pi[\tau\,|\, s_0=s]$. Hence,
\begin{align}
	\begin{split}\label{eq:firstprocess}
		\mathbb E_\pi[\tau\,|\, s_0=s,a_0=a]
		&=
		\mathbb E_{\pi'}[\sigma\,|\, s_0=s]\\
		&\quad
		+p\,\mathbb E_\pi[\tau\,|\, s_0=s]
		+(1-p)\cdot 0\\
		&=
		\mathbb E_{\pi'}[\sigma\,|\, s_0=s]
		+
		p\,\mathbb E_\pi[\tau\,|\, s_0=s].    
	\end{split}
\end{align}
Similarly, for the second process, if $s_\sigma=z$, the remaining absorption time is zero; if $s_\sigma=s$, then the process continues under the same stationary policy $\pi'$ from state $s$. Therefore, again by the strong Markov property,
\begin{align}\label{eq:secondprocess}
	\begin{split}
		\mathbb E_{\pi'}[\tau\,|\, s_0=s]
		&=
		\mathbb E_{\pi'}[\sigma\,|\, s_0=s]\\
		&\quad
		+p\,\mathbb E_{\pi'}[\tau\,|\, s_0=s]
		+(1-p)\cdot0\\
		&=
		\mathbb E_{\pi'}[\sigma\,|\, s_0=s]
		+
		p\,\mathbb E_{\pi'}[\tau\,|\, s_0=s].
	\end{split}
\end{align}
Eliminating $\mathbb E_{\pi'}[\sigma\,|\, s_0=s]$ in~\eqref{eq:firstprocess} using~\eqref{eq:secondprocess} gives
\[
\begin{aligned}
	&\quad\mathbb E_\pi[\tau\,|\, s_0=s,a_0=a]
	\\&=
	(1-p)\mathbb E_{\pi'}[\tau\,|\, s_0=s]
	+
	p\,\mathbb E_\pi[\tau\,|\, s_0=s]\\
	&\leq \frac{m}{1-\alpha},
\end{aligned}
\]
where we used the uniform expected hitting-time bound~\eqref{eq:expectationtau}. 
For $s=z$, the same bound holds trivially. Hence $ \mathbb{E}_\pi[\tau \,|\, s_0=s,a_0=a] \leq \frac{m}{1-\alpha} $ holds for every state-action pair $(s,a)$ and every stationary policy $\pi$. Since $|R(s,a)|\leq R_{\max}$ and rewards vanish after absorption,
\begin{equation}
	\label{eq:boundedQ}
	\begin{aligned}
		|Q^\pi(s,a)|
		&=
		\left|
		\mathbb E_\pi
		\left[
		\sum_{t=0}^{\tau-1}\gamma^tR_t
		\, | \, 
		s_0=s,a_0=a
		\right]
		\right|\\
		&\leq
		R_{\max}
		\mathbb E_\pi
		\left[
		\sum_{t=0}^{\tau-1}\gamma^t
		\, | \, 
		s_0=s,a_0=a
		\right]\\
		&\leq
		R_{\max}
		\mathbb E_\pi[\tau\,|\, s_0=s,a_0=a]\\
		&\leq
		\frac{m R_{\max}}{1-\alpha}.
	\end{aligned}
\end{equation}
We now bound the gradient mismatch. In the episodic case,
\begin{equation*}\begin{adjustbox}{max width=\columnwidth}$\displaystyle
\nabla_\theta^{\textup{e}} J -\hat{\nabla}^{\textup{e}}_\theta J =
\sum_s
\bigl(d_{\pi,\gamma}(s)-d_\pi(s)\bigr)
\sum_a
\nabla_\theta\pi(a\,|\, s)Q^\pi(s,a).
$\end{adjustbox}\end{equation*}
Therefore, by the triangle inequality,
\[
\begin{aligned}
	&\quad \|\nabla_\theta^{\textup{e}} J -\hat{\nabla}^{\textup{e}}_\theta J \|_2\\
	&\leq
	\sum_s
	|d_{\pi,\gamma}(s)-d_\pi(s)|
	\sum_a
	\|\nabla_\theta\pi(a\,|\, s)\|_2
	|Q^\pi(s,a)|\\
	&\leq
	\frac{mR_{\max}G |\mathcal{A}|}{1-\alpha}    \left \|  \tilde{d}_{\pi,\gamma} - \tilde{d}_\pi \right\|_1,
\end{aligned}
\]
where we used $\|\nabla_\theta\pi(a\,|\, s)\|_2\leq G$ from \cref{assumption:parameterization} and the uniform bound on $Q^\pi$ in~\eqref{eq:boundedQ}. Finally, by~\cref{thm:boundforepisodic}, 
\[
\begin{aligned}
	\left \| \nabla_\theta^{\textup{e}} J - \hat{\nabla}^{\textup{e}}_\theta J \right \|_2 
	&\le
	\frac{G|\mathcal A|mR_{\max}}{1-\alpha}
	\frac{2m}{1-\alpha}(1-\gamma)\\
	&=
	\frac{2G|\mathcal A|m^2R_{\max}}{(1-\alpha)^2}(1-\gamma)\triangleq\varepsilon_{\mathrm e}(\gamma).
\end{aligned}
\]
\end{proof}

\subsection{\texorpdfstring{Proof of \cref{thm:gradbias_cont}}{Proof: thm gradbias cont}}
\label{proof:gradbias_cont}
\begin{proof}
In continuing MDPs, the state-action value function $Q^\pi$ may grow as $(1-\gamma)^{-1}$ when $\gamma\to1$. We therefore exploit the fact that an action-independent baseline does not affect the policy gradient and rewrite the statewise gradient contribution in terms of the advantage function $A^\pi$. 
For a fixed policy $\pi$, define $g^\pi(s) : = \sum_{a \in \mathcal{A}} \nabla_\theta \pi(a   \, | \,  s) Q^\pi(s, a)$. Since $\sum_{a \in \mathcal{A}} \nabla_\theta \pi(a   \, | \,  s)=\nabla_\theta \sum_{a \in \mathcal{A}} \pi(a   \, | \,  s)=\nabla_\theta 1=0$,
we may subtract the value function $V^\pi(s)$ inside the summation and obtain
\begin{equation}\label{eq:adv_gradient_representation}
	\begin{aligned}
		g^\pi(s)&=\sum_{a \in \mathcal{A}} \nabla_\theta \pi(a   \, | \,  s)\left(Q^\pi(s, a) -V^\pi(s) \right) \\
		& = \sum_{a \in \mathcal{A}} \nabla_\theta \pi(a   \, | \,  s) A^\pi(s, a).
	\end{aligned}
\end{equation}
We next derive a uniform bound on $A^\pi$. Define the expected one-step reward under $\pi$ by ${r}_\pi(s) = \sum_{a\in\mathcal{A}} \pi(a   \, | \,  s) R(s, a)$ and its stationary mean by  $\eta_\pi:=\sum_{s \in \mathcal{S}} d_\pi (s) {r}_\pi(s).$ Let $ \bar{r}_\pi(s):=r_\pi(s)-\eta_\pi.
$ Then $ \sum_{s \in \mathcal{S}} d_\pi (s) \bar{r}_\pi(s)=0.
$ Using $ V^\pi=\left(I-\gamma P^\pi\right)^{-1} r_\pi $, $r_\pi=\eta_\pi \text{\usefont{U}{bbold}{m}{n}1}+\bar{r}_\pi$ and $P^\pi\text{\usefont{U}{bbold}{m}{n}1}=\text{\usefont{U}{bbold}{m}{n}1}$, we obtain
\begin{equation}
	\label{eq:VpiHpi}
	V^\pi
	=
	\frac{\eta_\pi}{1-\gamma}\text{\usefont{U}{bbold}{m}{n}1}
	+
	H_{\pi,\gamma},
\end{equation}
where $H_{\pi,\gamma}:= (I-\gamma P^\pi)^{-1}\bar r_\pi =\sum_{t=0}^{\infty} \gamma^t(P^\pi)^t\bar r_\pi $. 
Since $ d_\pi ^\top \bar{r}_\pi=0$, %
$ \left(P^\pi\right)^t \bar{r}_\pi=\left(\left(P^\pi\right)^t- \text{\usefont{U}{bbold}{m}{n}1}d_\pi ^\top\right) \bar{r}_\pi .
$ Therefore, 
$$
H_{\pi, \gamma}=\sum_{t=0}^{\infty} \gamma^t\left(\left(P^\pi\right)^t-\text{\usefont{U}{bbold}{m}{n}1}d_\pi ^\top\right) \bar{r}_\pi .
$$
By the geometric convergence result in {\eqref{eq:geo}}, for every state $s$, $\|(P^\pi)^t(s,\cdot)-d_\pi\|_1 \leq 2D\beta^t$. Hence, 
\[
\begin{aligned}
	\|H_{\pi,\gamma}\|_\infty
	&\leq
	\sum_{t=0}^{\infty}
	\gamma^t
	\max_{s\in\mathcal S}
	\|(P^\pi)^t(s,\cdot)-d_\pi\|_1
	\|\bar r_\pi\|_\infty\\
	&\leq
	2D\|\bar r_\pi\|_\infty
	\sum_{t=0}^{\infty}(\gamma\beta)^t\\
	&=
	\frac{2D}{1-\gamma\beta}
	\|\bar r_\pi\|_\infty.
\end{aligned}
\]
Since $|R(s,a)|\le R_{\max}$,  $|r_\pi(s)|\leq \sum_{a \in \mathcal{A}} \pi(a   \, | \,  s) R_{\max} = R_{\max} $ and $\left|\eta_\pi\right|=\left|\sum_{s \in \mathcal{S}} d_\pi(s) r_\pi(s)\right|\leq  R_{\max}$,  and hence $$\left\|\bar{r}_\pi\right\|_{\infty} \leq  \|r_\pi\|_\infty + |\eta_\pi|  \leq 2 R_{\max }.$$
Thus, 
\begin{equation}\label{eq:Hbound}
	\left\|H_{\pi, \gamma}\right\|_{\infty} \leq \frac{4 D R_{\max }}{1-\gamma\beta}.
\end{equation}
We now bound the advantage function. From the Bellman equations $ Q^\pi(s, a)=R(s, a)+\gamma \sum_{s^{\prime}} \mathbb{P}\left(s^{\prime}   \, | \,  s, a\right) V^\pi\left(s^{\prime}\right) $ and $ V^\pi(s)=r_\pi(s)+\gamma \sum_{s^{\prime}} P^\pi\left(s, s^{\prime}\right) V^\pi\left(s^{\prime}\right) $, we obtain
\[
\begin{aligned}
	A^\pi(s, a)&=R(s, a)-r_\pi(s) \\
	&\quad +\gamma \sum_{s^{\prime}}\left(\mathbb{P}\left(s^{\prime}   \, | \,  s, a\right)-P^\pi\left(s, s^{\prime}\right)\right) V^\pi\left(s^{\prime}\right) .
\end{aligned}
\]
Substituting~\eqref{eq:VpiHpi}, the constant component $\eta_\pi/(1-\gamma)$ vanishes because $\sum_{s'} \bigl( \mathbb P(s'  \, | \,  s,a)-P^\pi(s,s') \bigr) = 0$. Therefore,
\[
\begin{aligned}
	A^\pi(s, a)&=  R(s, a)-r_\pi(s) \\
	&\quad +\gamma \sum_{s^{\prime}}\left(\mathbb{P}\left(s^{\prime}   \, | \,  s, a\right)-P^\pi\left(s, s^{\prime}\right)\right)H_{\pi, \gamma}\left(s^{\prime}\right), 
\end{aligned}
\]
and using~\eqref{eq:Hbound}, we obtain
\begin{equation}\label{eq:advantage_uniform_bound}
	\begin{aligned}
		\left| A^\pi(s, a) \right|  &\leq \left| R(s, a)-r_\pi(s) \right| \\
		&\quad +\gamma \sum_{s^{\prime}}\left|\mathbb{P}\left(s^{\prime}   \, | \,  s, a\right)-P^\pi\left(s, s^{\prime}\right)\right| \left | H_{\pi, \gamma}\left(s^{\prime}\right) \right | \\
		& \leq 2R_{\max} + 2 \gamma \left \| H_{\pi, \gamma} \right \|_\infty \\
		& \leq 2R_{\max} + \frac{8\gamma D R_{\max }}{1-\gamma\beta}.
	\end{aligned}
\end{equation}
Combining \eqref{eq:adv_gradient_representation}, \eqref{eq:advantage_uniform_bound}, and $\|\nabla_\theta\pi(a  \, | \,  s)\|_2\leq G$, we have
$$
\begin{aligned}
	\|g^\pi(s)\|_2  &  \leq \sum_{a\in \mathcal{A}} \| \nabla_\theta \pi(a   \, | \,  s)\|_2 \cdot \left| A^\pi(s, a) \right|  \\
	& \leq G|\mathcal{A}| \left(2 R_{\max }+\frac{8\gamma D R_{\max }}{1-\gamma\beta}\right).
\end{aligned}
$$
Finally,
\[
\begin{aligned}
	&\quad  \left \| \nabla_\theta^{\textup{e}} J - \hat{\nabla}^{\textup{e}}_\theta J \right\|_2 \\
	& =\left \| \sum_s \left( d_{\pi,\gamma} (s) - d_\pi (s) \right) \sum_a \nabla_\theta \pi(a   \, | \,  s) Q^\pi(s, a)\right\|_2 \\
	& =\left \| \sum_s \left( d_{\pi,\gamma} (s) - d_\pi (s) \right) g^\pi (s) \right\|_2 \\
	& \leq \sum_{s\in \mathcal{S}} \left| d_{\pi,\gamma} (s) - d_\pi (s) \right| \cdot \|g^\pi(s) \|_2 \\
	& \leq G|\mathcal{A}| \left(2 R_{\max }+\frac{8\gamma D R_{\max }}{1-\gamma\beta}\right) \|d_{\pi,\gamma} -d_\pi \|_1 .
\end{aligned}
\]
Invoking \cref{thm:boundforcontinuing},
\begin{equation*}\begin{adjustbox}{max width=\columnwidth}$\displaystyle
\begin{aligned}
	\left \| \nabla_\theta^{\textup{e}} J - \hat{\nabla}^{\textup{e}}_\theta J \right\| \\
	\leq G|\mathcal A|\left(2R_{\max}+\frac{8\gamma DR_{\max}}{1-\gamma\beta}\right) \frac{2 D}{1-\beta \gamma}(1-\gamma)\triangleq \varepsilon_{\mathrm{c}}(\gamma).
\end{aligned}
$\end{adjustbox}\end{equation*}
\end{proof}

\section{\texorpdfstring{Proof of
	\cref{subsection:performance}}{Proof: Optimization implications}}
	\label{appendix:performance}
	\begin{proof}
We  give a unified proof for episodic and continuing MDPs. Define $\Delta_t:=\hat{\nabla}^{\textup{e}}_\theta J (\theta_t)-\nabla_\theta^{\textup{e}} J (\theta_t)$. Then the biased policy-gradient update in~\eqref{eq:biasedPG} can be written as $\theta_{t+1} =\theta_t+\eta_t\bigl(\nabla_\theta^{\textup{e}} J (\theta_t)+\Delta_t\bigr)$. By \cref{thm:gradbias_ep,thm:gradbias_cont}, we have $\|\Delta_t\|_2\leq\varepsilon_\gamma$ for all $t$, where $\varepsilon_\gamma=\varepsilon_{\mathrm e}(\gamma)$ in episodic MDPs and $\varepsilon_\gamma=\varepsilon_{\mathrm c}(\gamma)$ in continuing MDPs.

Recall from \eqref{eq:epJ} and \eqref{eq:kappa} that
\begin{equation}
    \label{eq:true_normalized_gradient_relation}
    \nabla_\theta J (\theta_t)
    =
    \kappa_{\pi_{\theta_t}}\nabla_\theta^{\textup{e}} J (\theta_t).
\end{equation}
Moreover, the time-zero contribution to the
discounted occupancy measure gives
\[
    \kappa_{\pi_{\theta_t}}
    =
    \sum_{s\in\tilde{\mathcal S} }
    \mu_{\pi_{\theta_t},\gamma}(s)
    \geq
    \sum_{s\in\tilde{\mathcal S} }d_0(s)
    =1.
\]

By~\cref{assumption:smoothness}, for any vector $u$, 
$J(\theta_t+u)
\geq
J(\theta_t)
+\langle\nabla_\theta J (\theta_t),u\rangle
-\frac{L}{2}\|u\|_2^2$.
Taking  $u=\eta_t\hat{\nabla}^{\textup{e}}_\theta J (\theta_t)$ and using~\eqref{eq:true_normalized_gradient_relation}, we obtain
\begin{equation*}
	\begin{aligned}
		&\quad J(\theta_{t+1})-J(\theta_t) \\
		&\geq 
		\eta_t
		\langle\nabla_\theta J (\theta_t),\hat{\nabla}^{\textup{e}}_\theta J (\theta_t)\rangle
		-\frac{L\eta_t^2}{2}
		\|\hat{\nabla}^{\textup{e}}_\theta J (\theta_t)\|_2^2\\
		&= 
		\eta_t\kappa_{\pi_{\theta_t}}
		\langle\nabla_\theta^{\textup{e}} J (\theta_t),\hat{\nabla}^{\textup{e}}_\theta J (\theta_t)\rangle
		-\frac{L\eta_t^2}{2}
		\|\hat{\nabla}^{\textup{e}}_\theta J (\theta_t)\|_2^2.
	\end{aligned}
\end{equation*}
Since 	$\Delta_t=\hat{\nabla}^{\textup{e}}_\theta J (\theta_t)-\nabla_\theta^{\textup{e}} J (\theta_t)$,
\begin{equation*}\begin{adjustbox}{max width=\columnwidth}$\displaystyle
\begin{aligned}
		2\langle\nabla_\theta^{\textup{e}} J (\theta_t),\hat{\nabla}^{\textup{e}}_\theta J (\theta_t)\rangle
		&=
		\|\nabla_\theta^{\textup{e}} J (\theta_t)\|_2^2
		+\|\hat{\nabla}^{\textup{e}}_\theta J (\theta_t)\|_2^2
		-\|\Delta_t\|_2^2.
	\end{aligned}
$\end{adjustbox}\end{equation*}
Substituting this identity gives
\begin{equation*}
	\begin{aligned}
		J(\theta_{t+1})-J(\theta_t)
		&\geq 
		\frac{\eta_t\kappa_{\pi_{\theta_t}}}{2}
		\bigl(
		\|\nabla_\theta^{\textup{e}} J (\theta_t)\|_2^2-\|\Delta_t\|_2^2
		\bigr)\\
		&\quad +
		\frac{\eta_t}{2}
		\bigl(\kappa_{\pi_{\theta_t}}-L\eta_t\bigr)
		\|\hat{\nabla}^{\textup{e}}_\theta J (\theta_t)\|_2^2.
	\end{aligned}
\end{equation*}
Since $\kappa_{\pi_{\theta_t}}\geq1$ and
$0<\eta_t\leq1/L$, 	$\kappa_{\pi_{\theta_t}}-L\eta_t\geq 0$.
Dropping the resulting nonnegative term and using $\|\Delta_t\|_2\leq\varepsilon_\gamma$, we obtain
\begin{equation}\label{eq:sumbasis}
	\begin{aligned}
		J(\theta_{t+1})-J(\theta_t)\geq
		\frac{\eta_t\kappa_{\pi_{\theta_t}}}{2}
		\bigl(
		\|\nabla_\theta^{\textup{e}} J (\theta_t)\|_2^2-\varepsilon_\gamma^2
		\bigr).
	\end{aligned}
\end{equation}
Summing \eqref{eq:sumbasis} from $t=N$ to $T$ gives
\begin{multline*}
	J(\theta_{T+1})-J(\theta_N)\\\geq
	\frac12\sum_{t=N}^{T}
	\eta_t\kappa_{\pi_{\theta_t}}
	\|\nabla_\theta^{\textup{e}} J (\theta_t)\|_2^2-
	\frac{\varepsilon_\gamma^2}{2}
	\sum_{t=N}^{T}\eta_t\kappa_{\pi_{\theta_t}}.
\end{multline*}
Since $J(\theta_{T+1})\leq J^*$, it follows that
\begin{equation*}
	\begin{aligned}
		&\frac{
			\sum_{t=N}^{T}\eta_t\kappa_{\pi_{\theta_t}}
			\|\nabla_\theta^{\textup{e}} J (\theta_t)\|_2^2
		}{
			\sum_{t=N}^{T}\eta_t\kappa_{\pi_{\theta_t}}
		}\leq
		\frac{
			2\bigl(J^*-J(\theta_N)\bigr)
		}{
			\sum_{t=N}^{T}\eta_t\kappa_{\pi_{\theta_t}}
		}
		+\varepsilon_\gamma^2.
	\end{aligned}
\end{equation*}
Because all weights $\eta_t\kappa_{\pi_{\theta_t}}$ are positive, 
\begin{equation*}
	\begin{aligned}
		&\min_{N\leq t\leq T}
		\|\nabla_\theta^{\textup{e}} J (\theta_t)\|_2^2\leq
		\frac{
			2\bigl(J^*-J(\theta_N)\bigr)
		}{
			\sum_{t=N}^{T}\eta_t\kappa_{\pi_{\theta_t}}
		}
		+\varepsilon_\gamma^2.
	\end{aligned}
\end{equation*}
\begin{sloppypar}
Moreover, by 
$\kappa_{\pi_{\theta_t}}\geq1$ and
$\sum_{t=0}^{\infty}\eta_t=\infty$, $
\sum_{t=N}^{T}\eta_t\kappa_{\pi_{\theta_t}}
\geq
\sum_{t=N}^{T}\eta_t  \longrightarrow\infty$
as $T\to\infty$ for every fixed $N$.
Since bounded rewards imply
$J^*-J(\theta_N)<\infty$, letting  $T\to\infty$ yields
$\inf_{t\geq N}\|\nabla_\theta^{\textup{e}} J (\theta_t)\|_2^2
\leq\varepsilon_\gamma^2$.
Finally, letting $N\to\infty$ and using $\liminf_{t\to\infty}x_t
=
\lim_{N\to\infty}\inf_{t\geq N}x_t$,  we conclude that $	\liminf_{t\to\infty}\|\nabla_\theta^{\textup{e}} J (\theta_t)\|_2^2
	\leq\varepsilon_\gamma^2$. By \cref{thm:gradbias_ep,thm:gradbias_cont},
$\varepsilon_{\mathrm e}(\gamma)=O(1-\gamma)$ and
$\varepsilon_{\mathrm c}(\gamma)=O(1-\gamma)$. Therefore, in both episodic and continuing MDPs, $\liminf_{t\to\infty}
\|\nabla_\theta^{\textup{e}} J (\theta_t)\|_2^2
\leq
O\bigl((1-\gamma)^2\bigr)$. 

\end{sloppypar}
\end{proof}

\section{Experimental Configurations}
\label{app:experimental_configurations}

\subsection{Two-State Continuing MDP}
We consider a continuing MDP with two states $\mathcal{S}=\{s_1,s_2\}$ and two actions
$\mathcal{A}=\{a_1,a_2\}$. The transition probabilities are independent of the selected action, with transition matrix
\begin{equation*}
P=
\begin{bmatrix}
0.95 & 0.05\\
0.20 & 0.80
\end{bmatrix}.
\end{equation*}
The rewards are given by $R(s,a_1)=0$ and $R(s,a_2)=1$ for both states. Consequently, $a_2$ is the unique optimal action at each state, and $J^\star=1/(1-\gamma)$.

We set $\gamma=0.7$, $\eta=1$, and initialize all policy logits to zero. We consider two initial state distributions: $d_0^A=(0.99,0.01)$ and $d_0^B=(0.01,0.99)$. The stationary distribution is $d_\pi=(0.8,0.2)$ for any policy  $\pi$. The resulting theoretical asymptotic error ratios $C_{\mathrm{b}} / C_{\mathrm{e}}$ are $0.775$ and $1.936$ for Cases A and B, respectively.

\subsection{Stochastic Episodic Gridworld}
We consider the stochastic $4\times4$ gridworld based on \citep[Example 4.1]{RS-AB:2018}, as described in Section~\ref{subsection:tabular}. To ensure a unique optimal action at each nonterminal state, we introduce an action-dependent cost and 
$R(s,a)=-1-c(a)$, where $c(\text{up})=0$, $c(\text{down})=0.1$, $c(\text{left})=0.2$, and $c(\text{right})=0.3$.
We set $\gamma=0.7$, $\eta=1$, and initialize all policy logits to zero. Let $u$ denote the uniform distribution over the $14$ nonterminal states, and $e_i$ the distribution concentrated at state $i$, with states indexed from $0$ to $15$ in row-major order. The initial distributions for the two cases are $d_0^A=0.1u+0.9e_4$ and $d_0^B=0.1u+0.9e_3$. The corresponding theoretical asymptotic error ratios $C_{\mathrm{b}} / C_{\mathrm{e}}$  are $0.791$ and $1.083$, respectively.
Unlike the two-state MDP, the transition matrix in this gridworld depends on the policy. Consequently, both the discounted and undiscounted state distributions evolve during policy optimization.

\section*{Declaration of generative AI and AI-assisted
technologies in the manuscript preparation process}

During the preparation of this work the authors used ChatGPT  in order to improve the readability of the paper. After using this tool, the authors reviewed and edited the content as needed and take full responsibility for the content of the publication.

\bibliographystyle{elsarticle-harv}
\bibliography{wwzbib}

\end{document}